\pdfoutput=1
\documentclass[final,1p,times,authoryear]{elsarticle}

\usepackage{amssymb,amsthm,amsmath}
\usepackage{array}
\usepackage{natbib}
\usepackage{hyperref}
\usepackage{xcolor}
\usepackage{graphicx}
\usepackage[ruled,vlined]{algorithm2e}

\newcommand{\argmax}{\mathop{\mathrm{arg\,max}}}
\newcommand{\argmin}{\mathop{\mathrm{arg\,min}}}

\newcommand{\Rset}{\mathbb{R}}
\newcommand{\norm}[2][2]{\left\|#2\right\|_{#1}}
\newcommand{\uball}[1][]{{\mathcal{B}}_{#1}}

\newcommand{\group}[1][k]{{\mathcal G}_{#1}}

\newcommand{\bfe}{\mathbf{e}}
\newcommand{\bfg}{\mathbf{g}}

\newcommand{\bfy}{\mathbf{y}}

\newcommand{\bfX}{\mathbf{X}}

\newcommand{\bfbeta}{\boldsymbol\beta}

\newcommand{\bfgamma}{{\boldsymbol\gamma}}

\newcommand{\clD}{\mathcal{D}}

\newcommand{\clS}{\mathcal{S}}

\newcommand{\supp}{\mathcal{A}}

\newcommand{\hatbgamma}{\,\hat{\!\bfgamma}}

\newcommand{\hatbbeta}{\,\hat{\!\bfbeta}}

\newcommand\mytexttt[1]{\texttt{\small #1}}

\newtheorem{lemma}{Lemma}
\newtheorem{proposition}{Proposition}
\newcommand\xylabellarge[4]{
  \setlength{\unitlength}{13cm}%
  \begin{picture}(1,0.6)%
    \put(0.04,0.025){\includegraphics[width=0.95\unitlength]{#1}}
    \put(0.54,0){\makebox[0cm]{\small#2}}
    \put(0,0.3){\rotatebox{90.0}{\makebox[0cm]{\small#3}}}
    \put(0.525,0.6){\makebox[0cm]{#4}}
  \end{picture}%
}
\newcommand\xylabelsquare[4]{
  \setlength{\unitlength}{6.5cm}%
  \begin{picture}(1,1)%
    \put(0.075,0.075){\includegraphics[width=0.90\unitlength]{#1}}
    \put(0.575,0){\makebox[0cm]{\small#2}}
    \put(0,0.525){\rotatebox{90.0}{\makebox[0cm]{\small#3}}}
    \put(0.525,1){\makebox[0cm]{#4}}
  \end{picture}%
}

\newcommand\smallxylabelsquare[4]{
  \setlength{\unitlength}{0.25\linewidth}%
  \begin{picture}(1,1)%
    \put(0.10,0.10){\includegraphics[width=0.80\unitlength]{#1}}
    \put(0.5,0){\makebox[0cm]{\small#2}}
    \put(0,0.5){\rotatebox{90.0}{\makebox[0cm]{\small#3}}}
    \put(0.525,0.95){\makebox[0cm]{#4}}
  \end{picture}%
}

\newif\ifverylong\verylongtrue
\newif\iflong\longtrue

\journal{arXiv}

\begin{document}

\begin{frontmatter}




\title{Sparsity by Worst-Case Penalties}

\author[label1]{Yves Grandvalet}
\ead{yves.grandvalet@utc.fr}
\author[label2]{Julien Chiquet}
\ead{julien.chiquet@inra.fr}
\author[label3]{Christophe Ambroise}
\ead{christophe.ambroise@genopole.cnrs.fr}
\address[label1]{Sorbonne universit\'es, Universit\'e de technologie de Compi\`egne, CNRS, \\ Heudiasyc UMR 7253, CS 60 319, 60 203 Compi\`egne cedex, France}
\address[label2]{AgroParisTech, INRA, Universit\'e Paris-Saclay\\UMR MIA-Paris, 16 rue Claude Bernard, 75005, Paris, France}
\address[label3]{Universit\'e d'\'Evry Val d'Essonne, Universit\'e Paris-Saclay,  ENSIIE, USC INRA\\ UMR CNRS 8071, LaMME, 91000 \'Evry, France}

\begin{abstract}%
  This paper proposes a new interpretation of sparse penalties such as
  the elastic-net and the group-lasso.  Beyond providing a new
  viewpoint on these penalization schemes, our approach results in a
  unified optimization strategy.  Our experiments demonstrate that
  this strategy, implemented on the elastic-net, is computationally
  extremely efficient for small to medium size problems.  Our
  accompanying software solves problems very accurately, at machine
  precision, in the time required to get a rough estimate with
  competing state-of-the-art algorithms.  We illustrate on real and
  artificial datasets that this accuracy is required to for the
  correctness of the support of the solution, which is an important
  element for the interpretability of sparsity-inducing penalties.
\end{abstract} 


\end{frontmatter}

\section{Introduction}

Inferential statistics aim at drawing conclusions from data and from some kind
of assumption or prior information about the underlying distribution.  It is
well known that processes where data guide the choice of assumptions can lead to
paradoxes, resulting from overfitting issues, in particular in the large
dimensional setup when data are used to select explicative variables: the
seeming explanatory power of weakly relevant variables can be important when the
number of variables is similar to the number of data
points~\citep{Freedman83b,Ambroise02}.
In this context, we present here an unusual reformulation of variable selection
methods based on sparsity-inducing penalties: we show that these methods can be
interpreted as adaptive penalties, where adaptivity refers to the choice of the
eventual penalty from data.
However, contrary to the available reformulations so far, ours shows that the adaptation
of the sparsity-inducing penalty to data follows a disagreement principle, where
the least-favorable penalty is selected from data.

We believe that our new viewpoint can be instrumental in unified analyses of
algorithms and their derived estimators, and we show here two such examples:
first, we provide a generic algorithm for solving the data fitting problem;
then, we derive the general form of an optimality gap that may be used to
monitor convergence.
Our experimental section illustrates the effectiveness of the generic algorithm
motivated by our interpretation, when instanced on the elastic-net estimator:
the algorithm, which relies on solving linear systems
is accurate, and computationally efficient up to medium scale problems (thousands of
variables).
As a side experimental result, we show that solving problems with high
precision, as with the proposed approach, benefits to the performances, either
measured in terms of prediction accuracy or in terms of support error rate.

\section{Adaptive Penalties \label{sec:adaptquadra}}

\subsection{Background}

We consider the linear regression model 
\begin{equation}
  \label{eq:linear_reg_group}
  Y = X \bfbeta^\star + \varepsilon
  \enspace,
\end{equation}
where $Y$ is a continuous response variable, $X=(X_1,\dots,X_p)$ is a vector of
$p$ predictor variables, $\bfbeta^\star$ is the vector of unknown parameters and
$\varepsilon$ is a zero-mean Gaussian error variable with variance $\sigma^2$.
We will assume throughout this paper that $\bfbeta^\star$ has few non-zero
coefficients.

The estimation and inference of $\bfbeta^\star$ is based on training data,
consisting of a vector
$\mathbf{y}=(y_1,\dots,y_n)^\intercal$ for responses and a
$n\times  p$ design  matrix $\mathbf{X}$  whose $j$th  column contains
$\mathbf{x}_j  = (x_j^1,\dots,x_j^n)^\intercal$, the  $n$ observations
for variable $X_j$.  For  clarity, we assume that both $\mathbf{y}$
and $\{\mathbf{x}_j\}_{j=1,\dots,p}$ are centered so as to eliminate the
intercept from fitting criteria.

Penalization methods that build on the $\ell_1$-norm, referred to as
\emph{Lasso} procedures (Least Absolute Shrinkage and Selection Operator), are
now widely used to tackle simultaneously variable estimation and selection in
sparse problems.  They define a shrinkage estimator of the form
\begin{equation}\label{eq:general:original}
  \hatbbeta = \argmin_{\bfbeta\in\mathbb{R}^p} 
    \frac{1}{2} \norm{\bfX \bfbeta - \bfy}^2 + 
    \lambda \norm[]{\bfbeta}
  \enspace, 
\end{equation}
where $\norm[2]{\cdot}$ is the Euclidean norm and $\norm[]{\cdot}$ is an
arbitrary norm, chosen to induce some assumed sparsity pattern (typically
$\ell_1$ or $\ell_{c,1}$ norms, where $c \in (1,\infty]$).
%

The existence of computationally efficient optimization procedures plays an
important role in the popularity of these methods.
Although various general-purpose convex optimization solvers could be used
\citep{boyd2004convex}, exploiting the structure of the regularization problem -- 
 especially the sparsity of solutions -- is essential in terms of computational
efficiency.
\citet{2012_FML_Bach} provided an overview of the families of techniques 
specifically  designed for solving this type of problems:
proximal methods, coordinate descent algorithms, reweighted-$\ell_{2}$
algorithms, working-set methods.
Stochastic gradient methods \citep{moulines2011non}, the
Frank-Wolfe algorithm \citep{lacoste2012block} or ADMM \citep[Alternating Direction
Method of Multipliers,][]{boyd2011distributed} have also recently gained in
popularity to the resolution of sparse problems.

We present below a new formulation of
Problem~\eqref{eq:general:original} that motivates an algorithm that
may seem reminiscent of reweighted-$\ell_{2}$ algorithms, but which is
in fact more closely related to working-set methods.
As for reweighted-$\ell_{2}$ algorithms, our proposal is based on the
reformulation of the sparsity-inducing penalty in terms of penalties
that are simpler to handle (linear or quadratic).  However, whereas
reweighted-$\ell_{2}$ algorithms rely on a variational formulation of
the sparsity-inducing norm that ends up in an augmented minimization
problem, our proposal is rooted in the duality principle, eventually
leading to a minimax problem that lends itself to a working-set
algorithm that will be presented in Section~\ref{sec:algo}.

\subsection{Dual Norms}

When the sparsity-inducing penalty is a norm, its sublevel sets can always be
defined as the intersection of linear or quadratic sublevel sets.  In other
words, if the optimization problem is written in the form of a constrained
optimization problem with inequality constraints pertaining to the penalty,
then, the feasible region can be defined as the intersection of linear or
quadratic regions. 
This fact, which is illustrated in Figures~\ref{fig:en-penalty} and
\ref{fig:group-penalty}, stems from the definition of dual norms:
\begin{equation*}
  \norm[]{\bfbeta} = \max_{\bfgamma\in\uball[*]^{1}} \bfgamma^\intercal \bfbeta
  \enspace,
\end{equation*}
where $\uball[*]^{1}$ is the unit ball centered at the origin defined from the dual
norm $\norm[*]{\cdot}$, \textit{i.e.}
$\uball[*]^{\eta}=\left\{\bfgamma\in\mathbb{R}^p:\norm[*]{\bfgamma}\leq \eta\right\}$.
Using this definition, Problem~\eqref{eq:general:original} can be reformulated
as
\begin{equation}\label{eq:general:primal}
  \hatbbeta = \argmin_{\bfbeta\in\mathbb{R}^p} 
  \max_{\bfgamma\in\uball[*]^{1}}
    \frac{1}{2} \norm{\bfX \bfbeta - \bfy}^2 + 
    \lambda \bfgamma^\intercal \bfbeta
  \enspace. 
\end{equation}
Technically, this formulation is the primal form of the original 
Problem~\eqref{eq:general:original} using the coupling function defined by the 
dual norm \citep[see e.g.][]{Gilbert16, Bonnans06}. 
It is interesting in the sense that the problem
\begin{equation*}
  \min_{\bfbeta\in\mathbb{R}^p} 
  \frac{1}{2} \norm{\bfX \bfbeta - \bfy}^2 + 
  \lambda \bfgamma^\intercal \bfbeta
\end{equation*}
is simple to solve for any value of $\bfgamma$, since it only requires solving 
a linear system. 
Indeed, the problem
\begin{eqnarray}
  \hatbgamma & = & \argmax_{\bfgamma\in\uball[*]^{1}}
    \frac{1}{2} \norm{\bfX \bfbeta - \bfy}^2 + 
    \lambda \bfgamma^\intercal \bfbeta \nonumber \\
     & = & \argmax_{\bfgamma\in\uball[*]^{1}}
    \bfgamma^\intercal \bfbeta \label{eq:optimal_gamma}
\end{eqnarray}
%
is usually straightforward to solve.
Besides the sparsity of $\hatbbeta$, the overall efficiency of our algorithm
relies also on the invariance of $\hatbgamma$ with respect to
large changes in $\bfbeta$. 
For the penalties we are interested in, $\hatbgamma$ takes its value in a
finite set, defined by the extreme points of the convex polytope $\uball[*]^{1}$.
This number of points typically increases exponentially in $p$, but, with the working-set
strategy, the number of configuration actually visited typically grows linearly
with the number of non-zero coefficients in the solution $\hatbbeta$.

\subsection{Relations with Other Methods}

The expansion in dual norm expressed in Problem \eqref{eq:general:primal} bears
some similarities with the first step of the derivation of very general duality
schemes, such as Fenchel's duality or Lagrangian duality.
It is however dedicated to the category of problems expressed as in
\eqref{eq:general:original}, thereby offering an interesting novel view of this
category of problems.
In particular, it provides geometrical insights on these methods and a generic
algorithm for computing solutions.  The associated algorithm, that relies on
solving linear systems is accurate, and efficient up to medium scale problems
(thousands of variables).

\subsection{Geometrical Interpretation}

\begin{figure}
  \begin{center} 
  \setlength{\unitlength}{0.33\linewidth}%
  \begin{picture}(1,1)%
    \put(0.075,0.075){\includegraphics[width=0.90\unitlength]{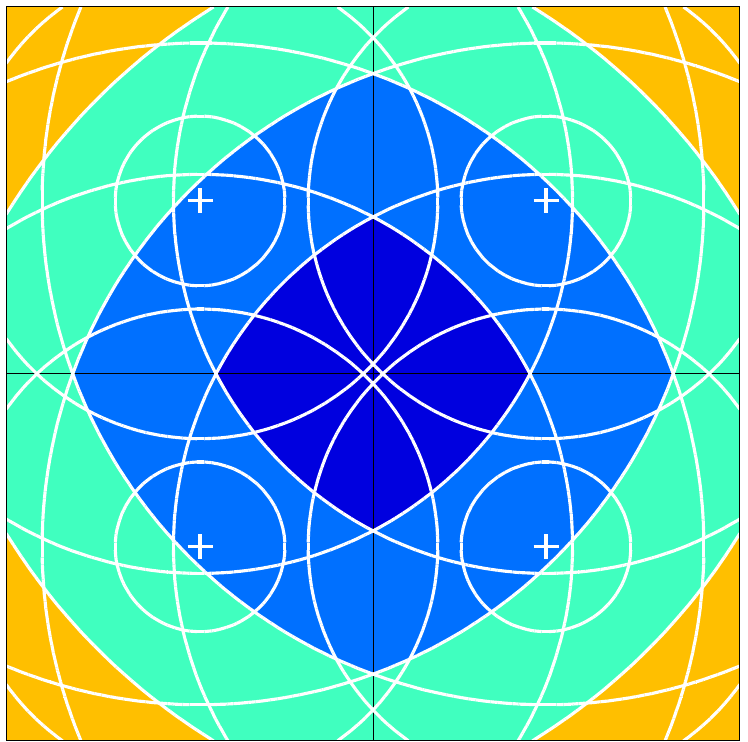}}
    \put(0.525,0){\makebox[0cm]{\small$\beta_1$}}
    \put(0,0.525){\rotatebox{90.0}{\makebox[0cm]{\small$\beta_2$}}}
    \put(0.525,1){\makebox[0cm]{Elastic Net}}
  \end{picture}%
  \setlength{\unitlength}{0.33\linewidth}%
  \begin{picture}(1,1)%
    \put(0.075,0.075){\includegraphics[width=0.90\unitlength]{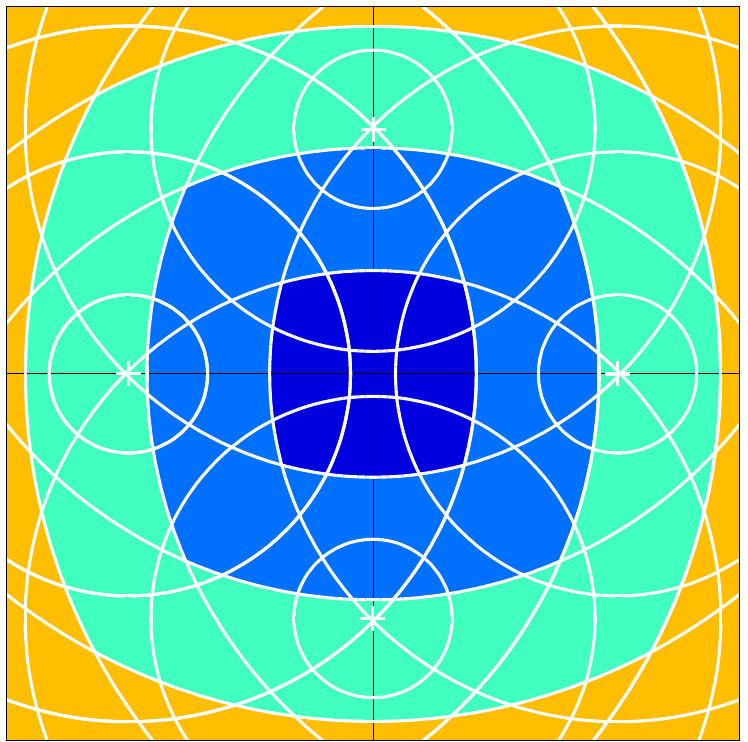}}
    \put(0.525,0){\makebox[0cm]{\small$\beta_1$}}
    \put(0,0.525){\rotatebox{90.0}{\makebox[0cm]{\small$\beta_2$}}}
    \put(0.525,1){\makebox[0cm]{$\ell_\infty$}}
  \end{picture}%
  \setlength{\unitlength}{0.33\linewidth}%
  \begin{picture}(1,1)%
    \put(0.075,0.075){\includegraphics[width=0.90\unitlength]{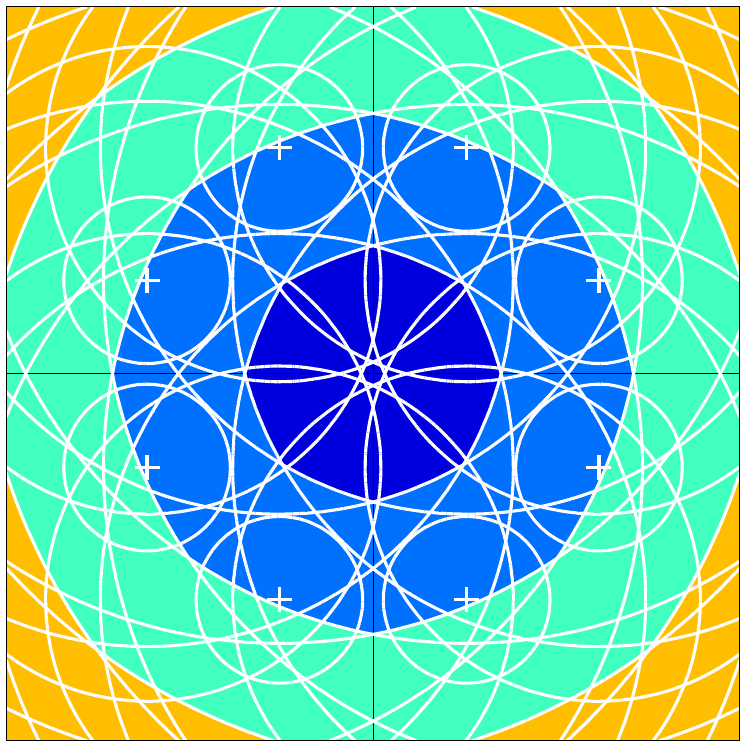}}
    \put(0.525,0){\makebox[0cm]{\small$\beta_1$}}
    \put(0,0.525){\rotatebox{90.0}{\makebox[0cm]{\small$\beta_2$}}}
    \put(0.525,1){\makebox[0cm]{OSCAR}}
  \end{picture}%
    \caption{Sublevel sets for several penalties (represented by the 
             colored patches).  
             Each set is defined as the intersection of the the Euclidean balls 
             whose boundaries are represented by white circles, and whose centers are
             represented by white crosses.}
     \label{fig:penalties}
    \end{center} 
\end{figure}
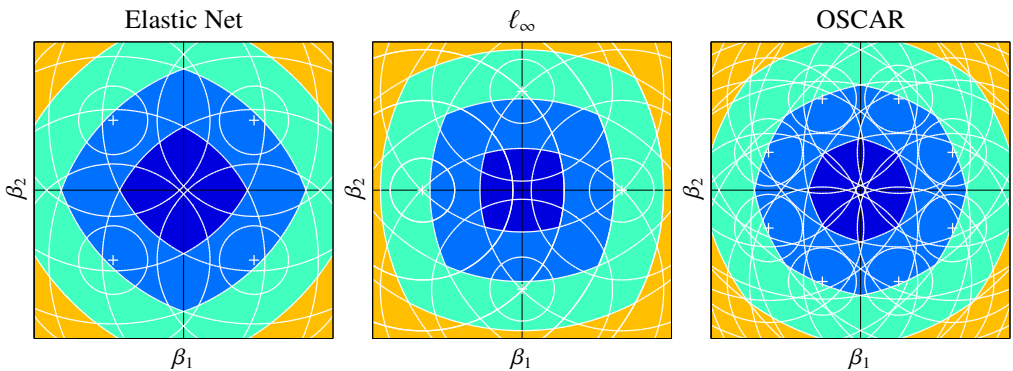

Geometrical insights are easier to gain from a slightly different formulation of
Problem \eqref{eq:general:primal}:~\footnote{%
  This quadratic formulation is more general, in the sense that it corresponds to
  Problem~\eqref{eq:general:primal} with a proper rescaling of $\lambda$ in the
  limit of $\eta\rightarrow+\infty$.
}
\begin{equation}\label{eq:general:dual}
  \min_{\bfbeta\in\mathbb{R}^p} \max_{\bfgamma\in\uball[*]^\eta}
    \frac{1}{2} \norm{\bfX \bfbeta - \bfy}^2 + \frac{\lambda}{2} \norm{\bfbeta - \bfgamma}^2
  \enspace.
\end{equation}
The penalty $\lambda\norm{\bfbeta - \bfgamma}^2$ corresponds to a hard
constraint $\norm{\bfbeta - \bfgamma}^2\leq c$, which states that the 
solution in $\bfbeta$ belongs to a $\ell_{2}$ ball centered in $\bfgamma$.
Then, as $\hatbgamma$ maximizes $\norm{\hatbbeta - \bfgamma}^2$, the
solution $\hatbbeta$ belongs to the intersection of all the balls
centered in $\bfgamma\in\uball[*]^\eta$.
Eventually, the active constraints will be defined by the $\bfgamma$ values for
which $\norm{\hatbbeta - {\bfgamma}^2}$ is maximal, that is for the
worst-case $\hatbgamma$ values. 
For the penalties we are interested in, $\hatbgamma$ takes its value in a
finite set, defined by the extreme points of the convex polytope $\uball[*]^\eta$.
%
%
This is the case for the Lasso,
the $\ell_{1,\infty}$ version of the group-lasso 
(where the magnitude of regression coefficients are assumed to be equal within
groups, either zero or non-zero), 
and for OSCAR (Octagonal Shrinkage and Clustering Algorithm for Regression)
which is based on a penalizer encouraging the sparsity of the regression
coefficients and the equality of the non-zero entries \citep{Bondell08}.
Figure \ref{fig:penalties} illustrates  those three  sparse problems
with their  associated worst case quadratic penalty.

%

\section{Worst-Case Quadratic Penalties \label{sec:quadra}}
\label{sec:gammaperturb}

Our framework is amenable to many variations.
Here, we present two examples where
the desired penalty on the regression coefficients $\bfbeta$ is implemented 
through the 
definition of the dual norm on the coefficients $\bfgamma$.
When the desired penalty on $\bfbeta$ is expressed by
$\ell_1$ or $\ell_\infty$ norms, this process results in unit balls
$\uball[*]^\eta$ which are convex polytopes that are 
easy to manage when solving Problem~\eqref{eq:general:dual}, since they
can be defined as the convex hulls of a finite number of values.

The two sparsity-inducing penalizers presented below have a grouping effect.
The elastic-net implements this grouping without predefining the group
structure: strongly correlated predictors tend to be in or out of the model
together \citep{2005_JRSS_Zou}.  
The $\ell_{\infty,1}$ group-Lasso that is presented subsequently is based on a
prescribed group structure and favors regression coefficients with identical
magnitude within activated groups.

\subsection{Elastic-Net} \label{sec:elasticnet}

\begin{figure}
  \begin{center} 
    \smallxylabelsquare{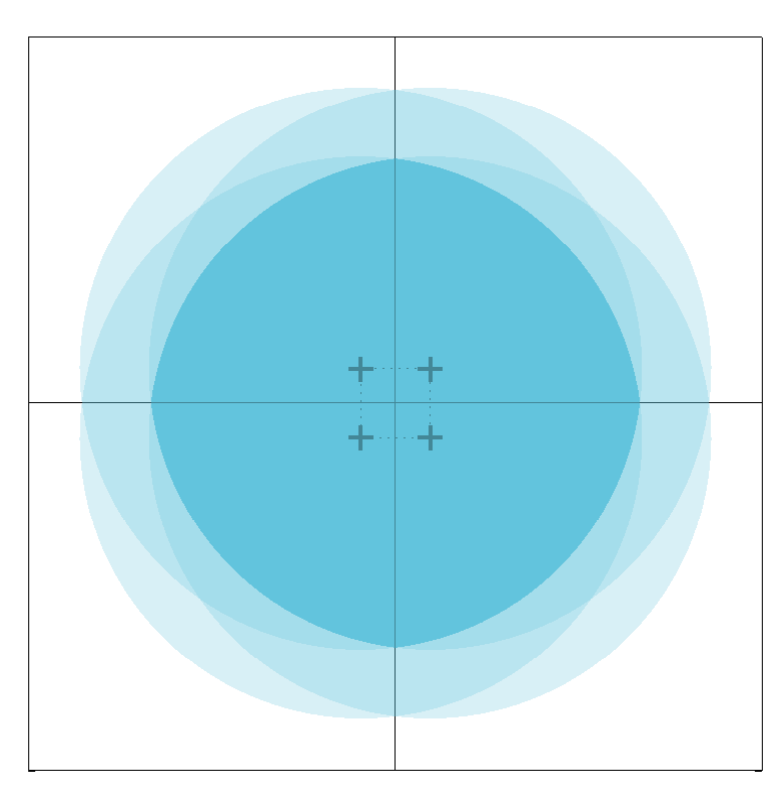}{$\beta_1$}{$\beta_2$}{}%
    \smallxylabelsquare{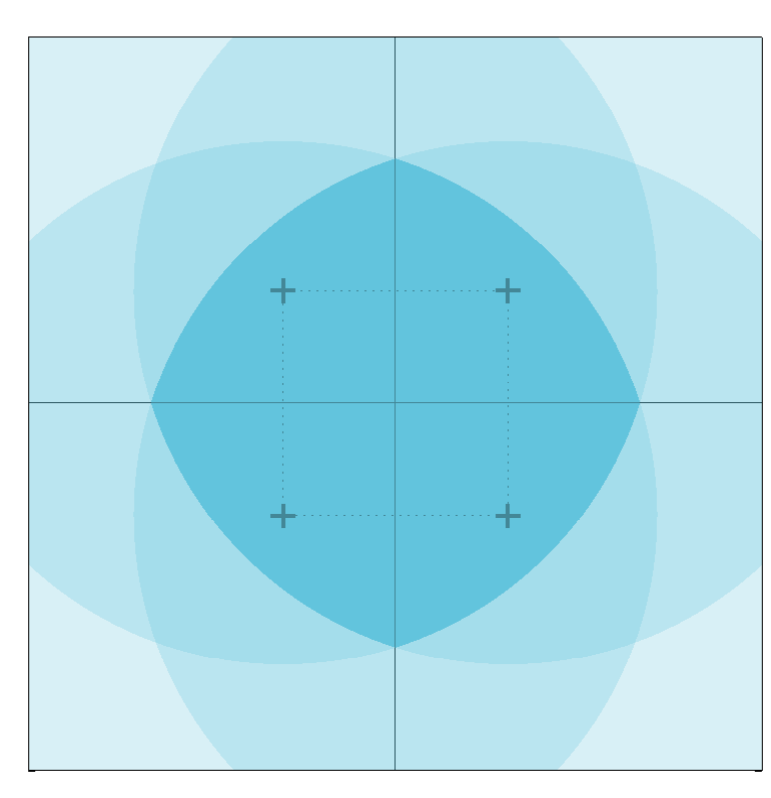}{$\beta_1$}{$\beta_2$}{}%
    \smallxylabelsquare{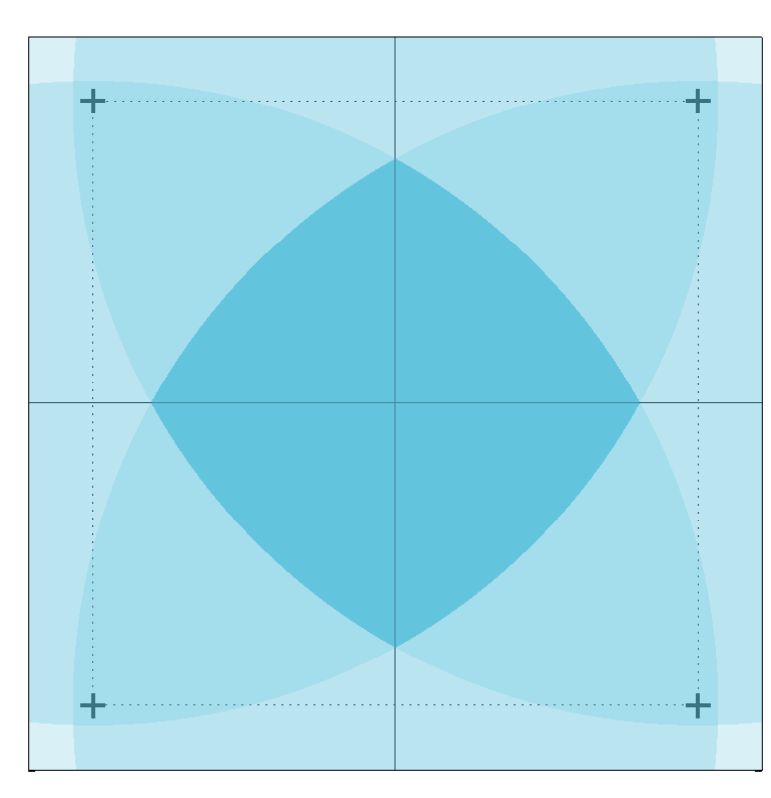}{$\beta_1$}{$\beta_2$}{}%
    \smallxylabelsquare{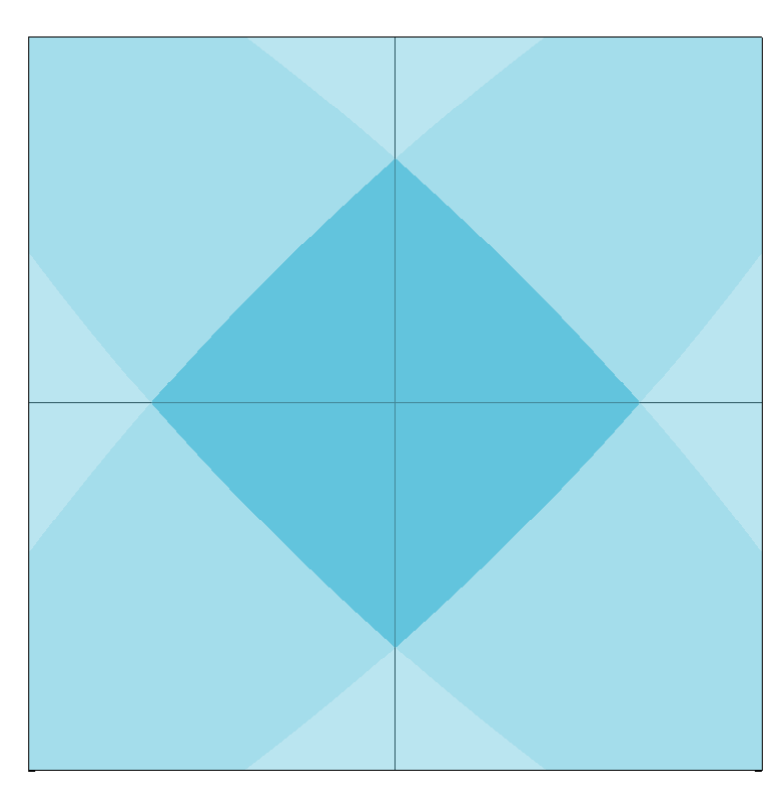}{$\beta_1$}{$\beta_2$}{}%
    \caption{Sublevel sets for elastic net penalties (represented by the darker
             colored patches).  
             Each set is defined as the intersection of the the Euclidean balls
             (represented by the lighter color patches) whose centers are
             represented by crosses.}
    \label{fig:en-penalty}
    \end{center} 
\end{figure}

As an introductory example, let us consider the assumption stating that the
$\ell_1$-norm of $\bfbeta^\star$ should be small. 
The dual norm is the $\ell_\infty$-norm:
\begin{align*}
  \uball[*]^\eta & = \left\{ \bfgamma \in \mathbb{R}^p :
\sup_{\norm[1]{\bfbeta}\leq1} \bfgamma^\intercal\bfbeta \leq \eta \right\} \\
    & = \left\{ \bfgamma \in \mathbb{R}^p : \norm[\infty]{\bfgamma} \leq \eta \right\} \\
    & = \mathbf{conv} \big\{ \left\{ -\eta, \eta \right\}^p \big\}
  \enspace,
\end{align*}
where $\mathbf{conv}$ denotes convex hull, so that Problem
\eqref{eq:general:dual} reads:
\begin{align}
  & \min_{\bfbeta\in\mathbb{R}^p} \max_{\bfgamma \in\uball[*]^\eta}
      \Big\{ \frac{1}{2} \norm{\bfX \bfbeta - \bfy}^2 + \frac{\lambda}{2} \norm{\bfbeta - \bfgamma}^2 
      \Big\} \nonumber \\
  \Leftrightarrow
    & \min_{\bfbeta\in\mathbb{R}^p}
       \frac{1}{2} \norm{\bfX \bfbeta - \bfy}^2 + \lambda \eta
       \norm[1]{\bfbeta} + \frac{\lambda}{2} \norm{\bfbeta}^2
  \enspace, \label{eq:elastic-net}
\end{align}
which is recognized as an elastic-net problem.
When $\eta$ is null, we recover ridge regression, and when $\eta$ goes to 
infinity, the problem approaches a Lasso problem.
A 2D pictorial illustration of this evolution is given in
Figure~\ref{fig:en-penalty}, where the shape of the uncertainty set
$\uball[*]^\eta$ is the convex hull of the points located at 
$(\pm \eta, \pm \eta)^\intercal$, which are identified by the cross markers.
Then, the sublevel set 
$\{\bfbeta : \max_{\bfgamma \in \uball[*]^\eta} \norm{\bfbeta-\bfgamma}^2 \leq t\}$
is simply defined as the intersection of the four sublevel sets
$\{\bfbeta : \norm{\bfbeta - \bfgamma}^2 \leq t\}$ for $\bfgamma=(\pm
1, \pm 1)^\intercal$, which are Euclidean balls centered at
these $\bfgamma$ values.

\subsection{Group-Lasso}

We consider here the $\ell_{\infty,1}$ variant of the group-Lasso, which was
first proposed by \citet{Turlach05} to perform variable selection in the
multiple response setup.
A group structure is defined on
the set of variables by setting a partition of the index set
$\mathcal{I}=\{1,\ldots,p\}$, that is,
$
  \mathcal{I}=\bigcup_{k=1}^K\group \enspace,\, \text{with}\enspace 
  \group \cap \group[\ell]=\emptyset \enspace
  \text{for}\enspace k\neq\ell \enspace.
$
Let $p_k$ denote the cardinality of group $k$, and $\bfbeta_{\group} \in
\Rset^{p_k}$ be the vector $(\beta_j)_{j\in \group}$.

The $\ell_{\infty,1}$ mixed-norm of $\bfbeta$ (that is, its groupwise 
$\ell_\infty$-norm) is defined as
\begin{equation*}
  \uball = \left\{ 
    \bfbeta \in \mathbb{R}^p :\sum_{k=1}^K \norm[\infty]{\bfbeta_{\group}} \leq 1
  \right\}
  \enspace.
\end{equation*}
The dual norm is the groupwise $\ell_1$-norm:
\begin{align*}
  \uball[*]^\eta & = \left\{ \bfgamma \in \mathbb{R}^p :
\sup_{\bfbeta\in\uball} \bfgamma^\intercal\bfbeta \leq \eta \right\} \\
    & = \left\{ \bfgamma \in \mathbb{R}^p : \max_{k\in\{1,...,K\}}  \norm[1]{\bfgamma_{\group}} \leq \eta \right\} \\
    & = \mathbf{conv} \big\{ 
                        \left\{\eta\bfe^{p_1}_1, \ldots, \eta\bfe^{p_1}_{p_1},-\eta\bfe^{p_1}_1, \ldots, -\eta\bfe^{p_1}_{p_1} \right\} 
                        \times \ldots \\
    & \hspace*{4em} \times 
                        \left\{\eta\bfe^{p_K}_1, \ldots, \eta\bfe^{p_K}_{p_K},-\eta\bfe^{p_K}_1, \ldots, -\eta\bfe^{p_K}_{p_K} \right\} 
                      \big\}
  \enspace,
\end{align*}
where $\bfe^p_j$ is the $j$th element of the canonical basis of $\Rset^p$.
Problem \eqref{eq:general:dual} becomes:
\begin{align*}
  & \min_{\bfbeta\in\mathbb{R}^p} \max_{\bfgamma \in \uball[*]^\eta}
      \Big\{ \frac{1}{2} \norm{\bfX \bfbeta - \bfy}^2 + \frac{\lambda}{2} \norm{\bfbeta - \bfgamma}^2 \Big\} \\
  \Leftrightarrow
    & \min_{\bfbeta\in\mathbb{R}^p}
      \frac{1}{2} \norm{\bfX \bfbeta - \bfy}^2 + \lambda \eta \sum_{k=1}^K \norm[\infty]{\bfbeta_{\group}} + \frac{\lambda}{2} \norm{\bfbeta}^2 
  \enspace,
\end{align*}

\begin{figure}
  \begin{center} 
    \smallxylabelsquare{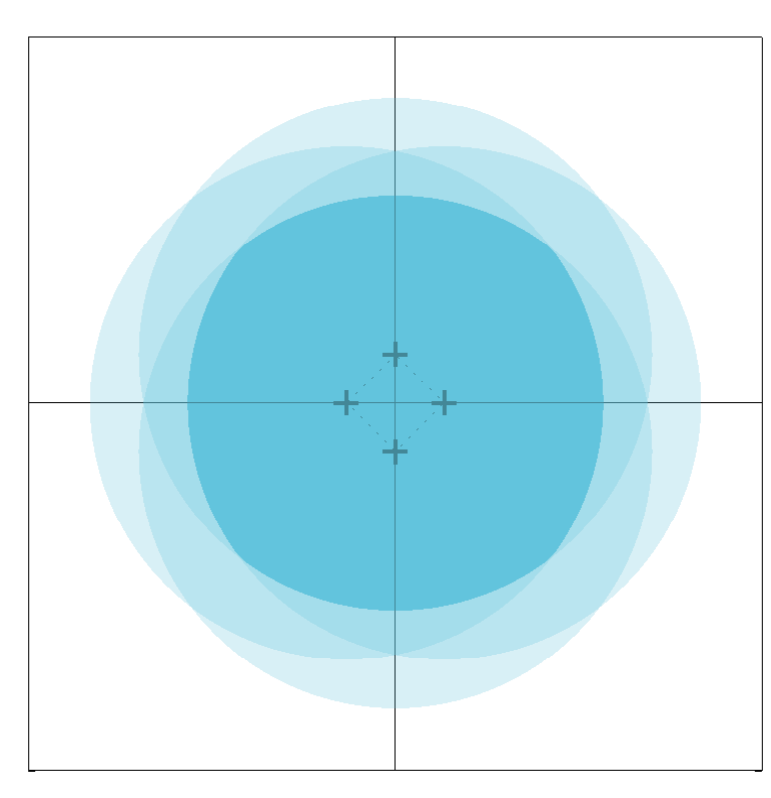}{$\beta_1$}{$\beta_2$}{}%
    \smallxylabelsquare{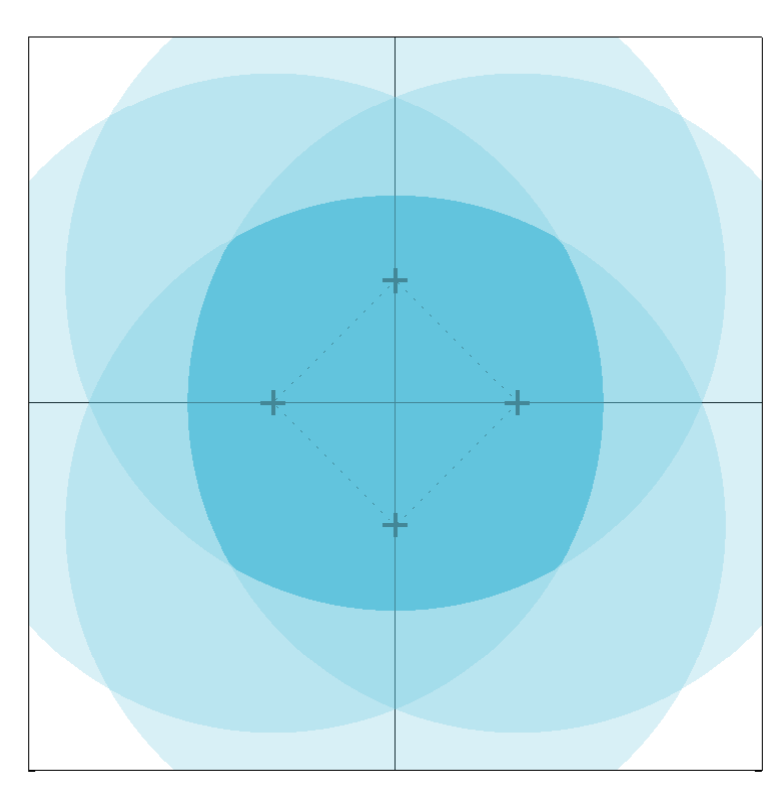}{$\beta_1$}{$\beta_2$}{}%
    \smallxylabelsquare{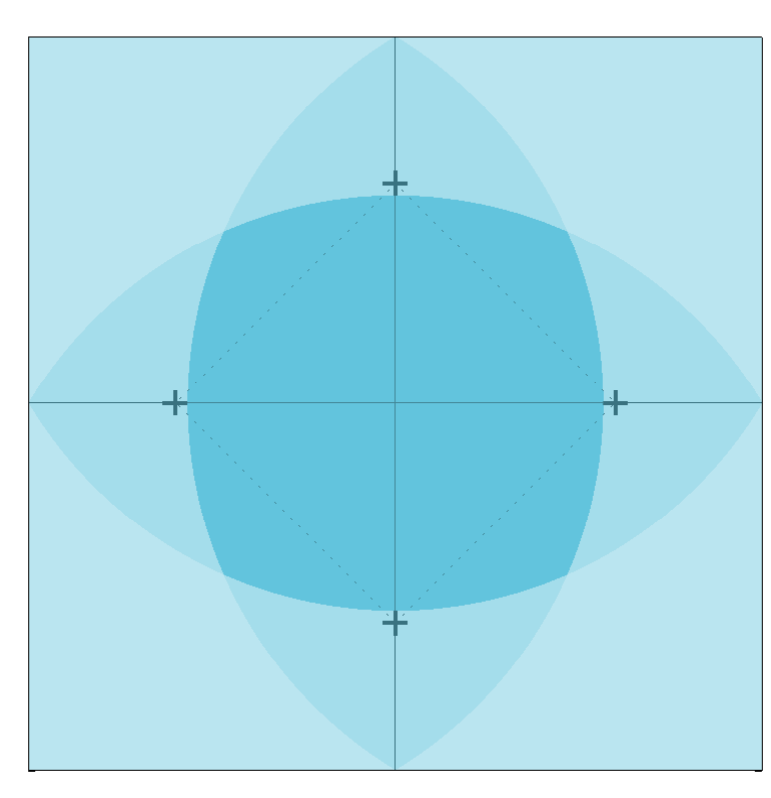}{$\beta_1$}{$\beta_2$}{}%
    \smallxylabelsquare{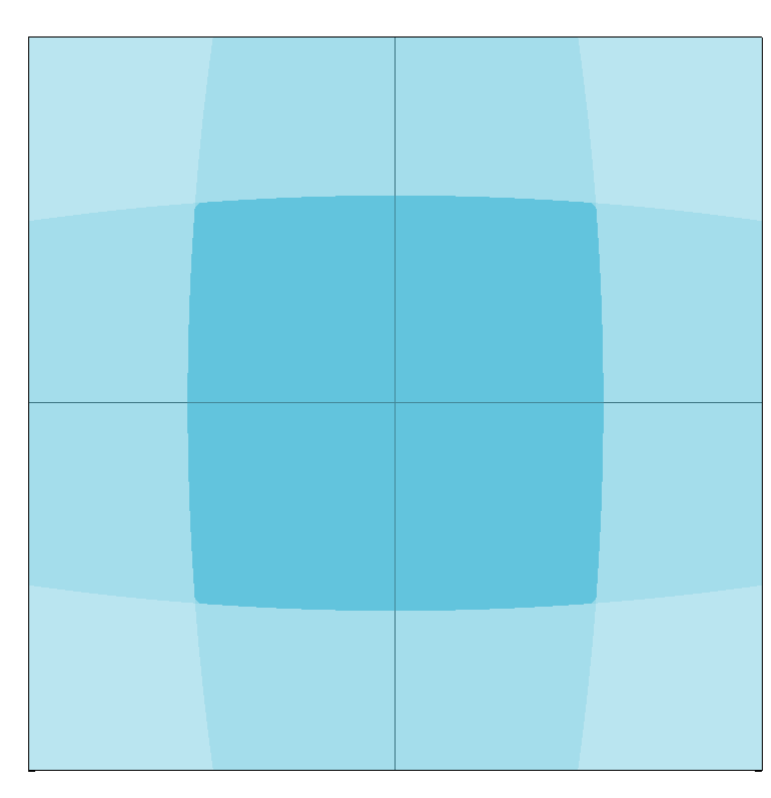}{$\beta_1$}{$\beta_2$}{}%
    \caption{Sublevel sets for the $\ell_{\infty,1}$ group-Lasso penalties
             (represented by the darker colored patches).
             Each set is defined as the intersection of the the Euclidean balls
             (represented by the lighter color patches) whose centers are
             represented by crosses.}
    \label{fig:group-penalty}
    \end{center} 
\end{figure}

Notice that the limiting cases of this penalty are two classical problems: ridge
regression and the $\ell_{\infty,1}$ group-Lasso.
A 2D pictorial illustration of this evolution is given in
Figure~\ref{fig:group-penalty}, where $\uball[*]^\eta$ is the convex hull of the
points located on the axes at $\pm \eta$, which are identified by the cross
markers.
Then, the sublevel set 
$\{\bfbeta : \max_{\bfgamma \in \uball[*]^\eta} \norm{\bfbeta-\bfgamma}^2 \leq t\}$
is simply defined as the intersection of the four sublevel sets
$\{\bfbeta : \norm{\bfbeta - \bfgamma}^2 \leq t\}$ for 
$\bfgamma=\pm \eta\bfe^{2}_1$ and $\bfgamma=\pm \eta\bfe^{2}_2$,
which are Euclidean balls centered at these $\bfgamma$ values.

\section{Algorithm}\label{sec:algo}

The unified derivation for the problems presented in Section
\ref{sec:gammaperturb} suggests a unified processing based on the iterative
resolution of quadratic problems. 
This general algorithm is summarized in this section.
We then show that the new derivation can also be used in the analysis of this 
algorithm by describing an alternative to Fenchel duality 
\citep[used for example by][]{2012_FML_Bach} to assess convergence.

\iflong
  \subsection{Active Set Approach}
\fi

The efficient approaches developed for sparse regression take advantage of the sparsity
of the solution by solving a series of small linear systems, the sizes of which are
incrementally increased/decreased.  
Here, as for the Lasso~\citep{2000_JCGS_Osborne,2004_AS_Efron}, this process boils down to an
iterative optimization scheme involving the resolution of quadratic problems.

The algorithm is based on the iterative update of the set of ``active''
variables, $\supp$, indexing the coefficients $\bfbeta_{\!\supp}$ currently
identified as being non-zero. 
It is started from a sparse initial guess, say $\supp=\emptyset$ ($\bfbeta=0$),
and iterates the three following steps:
\begin{enumerate}
\item \label{item:algo:step1} the first step solves
  Problem~\eqref{eq:general:dual} considering that $\supp$, the set of
  ``active'' variables, is correct; that is, the
  objective~\eqref{eq:general:dual} is optimized with respect to
  $\bfbeta_{\!\supp}$.  This penalized least squares problem is defined
  from $\mathbf{X}_{\centerdot\supp}$, which is the submatrix of
  $\mathbf{X}$ comprising all rows and the columns indexed by $\supp\,$
  and $\bfgamma_{\!\supp}$, which is set to
  its current most adversarial value.~\footnote{%
    When several $\bfgamma_{\!\supp}$ are equally unfavorable to
    $\bfbeta_{\!\supp}$, we use gradient information to pick the worst one
    among those when $\bfbeta_{\!\supp}$ moves along the steepest descent
    direction.
  }

\item \label{item:algo:step2} the second step updates $\bfbeta_{\!\supp}$
  if necessary (and possibly $\bfgamma_{\!\supp}$), so that
  $\bfgamma_{\!\supp}$ is indeed (one of) the most adversarial value of
  the current $\bfbeta_{\!\supp}$.
  This is easily checked with the problems given in
  Section~\ref{sec:gammaperturb}, where $\uball[*]^\eta$ is a convex polytope
  whose vertices (that is, extreme $\bfgamma$-values) are associated with a
  cone of coherent $\bfbeta$-value.

\item  \label{item:algo:step3} the last step updates the active set $\supp$. 
  It relies on the ``worst-case gradient'' with respect to
  $\bfbeta$, where $\bfgamma$ is chosen so as to minimize infinitesimal
  improvements of the current solution.
  Again picking the right $\bfgamma$ is easy for the problems given in
  Section~\ref{sec:gammaperturb}.
  Once this is done, we first check whether some variables should quit the
  active set, and if this is not the case, we assess the completeness of
  $\supp$, by checking the optimality conditions with respect to inactive
  variables.  We add the variable, or the group of variables that most violates
  the worst-case optimality conditions.  When no such violation exists, the
  current solution is optimal, since, at this stage, the problem is solved
  exactly within the active set $\supp$.
\end{enumerate}
Algorithm~\ref{algo:active_set} provides a more comprehensive technical
description.

\begin{algorithm}[htbp]
  \begin{small}
 \SetSideCommentLeft
  \nlset{Init.} $\bfbeta \leftarrow \bfbeta^0$\\
  Determine the active set: $\supp \leftarrow \{j:\beta_{j} > 0 \} $\\
  Pick a worst admissible $\bfgamma$, that is,  $\displaystyle\bfgamma \in 
  \argmax_{\bfgamma' \in \uball[*]^\eta} \left\| \bfgamma' - \bfbeta \right\|_2^2$
  \BlankLine 
  \nlset{Step 1} Update active variables $\bfbeta_{\!\supp}$ assuming that $\supp$ and $\bfgamma_{\!\supp}$ are optimal \\
  $\bfbeta_{\!\supp}^\mathrm{old} \leftarrow \bfbeta_{\!\supp}$ \\
  $\bfbeta_{\!\supp} \leftarrow 
   \left(\mathbf{X}_{\centerdot\supp}^\intercal
      \mathbf{X}_{\centerdot\supp} + \lambda \mathbf{I}_{|\supp|}\right)^{-1}
    \left(\mathbf{X}_{\centerdot\supp}^\intercal\mathbf{y} -
      \lambda \bfgamma_{\!\supp}\right)  
  $ \\ 
  \nlset{Step 2} Verify coherence of $\bfgamma_{\!\supp}$ with the updated $\bfbeta_{\!\supp}$\\
  \If(\tcp*[f]{if $\bfgamma_{\!\supp}$ is not worst-case})
   {$\displaystyle\norm{\bfbeta_{\!\supp} - \bfgamma_{\!\supp}}^2 < \max_{\bfg \in \uball[*]^\eta} \norm{\bfbeta_{\!\supp}-\bfg_{\!\supp}}^2$}
   {%
    \tcc{Backtrack towards the last $\bfgamma_{\!\supp}$-coherent solution:}
    $\bfbeta_{\!\supp} \leftarrow \bfbeta_{\!\supp}^\mathrm{old} + \rho (\bfbeta_{\!\supp} - \bfbeta_{\!\supp}^\mathrm{old})$ \\
    $\bfgamma_{\!\supp}$ is worst-case for $\bfbeta_{\!\supp}$, and there is another worst-case value $\widetilde\bfgamma_{\!\supp}$ \\
    \tcc{Check whether progress can be made with $\widetilde\bfgamma_{\!\supp}$}
    $\widetilde\bfbeta_{\!\supp} \leftarrow
    \left(\mathbf{X}_{\centerdot\supp}^\intercal
      \mathbf{X}_{\centerdot\supp} + \lambda \mathbf{I}_{|\supp|}\right)^{-1}
    \left(\mathbf{X}_{\centerdot\supp}^\intercal\mathbf{y} -
      \lambda \widetilde\bfgamma_{\!\supp}\right)  
  $ \\ 
  \If(\tcp*[f]{if $\widetilde\bfgamma_{\!\supp}$ is worst-case\ldots})
   {$\displaystyle\norm{\widetilde\bfbeta_{\!\supp} - \widetilde\bfgamma_{\!\supp}}^2 = \max_{\bfg \in \uball[*]^\eta} \norm{\widetilde\bfbeta_{\!\supp}-\bfg_{\!\supp}}^2$}
     {%
       $(\bfbeta_{\!\supp},\bfgamma_{\!\supp}) \leftarrow
          (\widetilde\bfbeta_{\!\supp},\widetilde\bfgamma_{\!\supp})$
        \tcp*[f]{$ (\widetilde\bfbeta_{\!\supp},\widetilde\bfgamma_{\!\supp})$ is better than $(\bfbeta_{\!\supp},\bfgamma_{\!\supp})$}
     }
    }
    \tcc{The current $\bfgamma_{\!\supp}$ is coherent with $\bfbeta_{\!\supp}$}

  \nlset{Step 3} Update active set $\supp$ \\
  $\displaystyle g_j \leftarrow \min_{\bfgamma\in\uball[*]^\eta} \left|
    \mathbf{x}_j^\intercal(\mathbf{X}_{\centerdot\supp}\bfbeta_{\!\supp} - \mathbf{y})  + \lambda (\beta_j - \gamma_j) \right|
  \enspace j=1,\ldots,p$
  \tcp*[f]{worst-case gradient} \\
  \eIf{$\exists\, j \in\supp:\beta_j = 0 \ \text{and}\ g_j = 0$}{
       $\displaystyle \supp \leftarrow \supp\backslash\{j\}$ \tcp*[r]{Downgrade $j$} 
       \tcc{Go to Step 1}
   }{%
  \eIf{$\max_{j\in\supp^c} g_j \neq0$}{ 
   \tcc{Identify the greatest violation of optimality conditions}
    $\displaystyle j^\star \leftarrow \argmax_{j\in\supp^c} g_j \,,\enspace$
    $\supp \leftarrow \supp \cup \{j^\star\}$ \tcp*[r]{Upgrade $j^\star$} 
    \tcc{Go to Step 1}
  }{
   \tcc{Stop and return $\bfbeta$, which is optimal}
   }
  }
  \end{small} 
\caption{Worst-Case Active Set Algorithm}
\label{algo:active_set}
\end{algorithm}
Note that the structure is essentially identical to the one proposed by
\citet{2000_JCGS_Osborne} or \cite{2004_AS_Efron} for the Lasso, but that it 
applies to any penalty that can be decomposed as in 
Problem~\eqref{eq:general:dual}.
Our viewpoint is also radically different, as the global non-smooth problem
is dealt with subdifferentials by \citet{2000_JCGS_Osborne}, whereas we rely on
the maximum of smooth functions.
This approach suggests a new assessment of convergence, as detailed below.
%

  \subsection{Monitoring Convergence}

  At each iteration of the algorithm, the current $\bfbeta$ is computed assuming
  that the current active set $\supp$ and the current
  $\bfgamma_\supp$-value are optimal.
  When the current active set is not optimal, the current $\bfbeta$ (where
  $\bfbeta_\supp$ is completed by zeros on the complement $\supp^c$) is
  nevertheless optimal for a $\bfgamma$-value defined in $\Rset^p$ (where
  $\bfgamma_\supp$ is completed by ad hoc values on the complement
  $\supp^c$). However this $\bfgamma$ fails to belong to $\uball[*]^\eta$ 
  (otherwise, the problem would be solved: $\supp$, $\bfgamma$ and $\bfbeta$
  would indeed be optimal).
  The following proposition relates the current objective function, associated
  with an infeasible $\bfgamma$-value ($\bfgamma\notin\uball[*]^\eta$), to the
  global optimum of the optimization problem.

  \begin{proposition}\label{prop:monitoring}
    For any vectorial norm $\norm[*]{\cdot}$, 
    $\forall \bfgamma \in \mathbb{R}^{p}:\norm[*]{\bfgamma} \geq \eta$, we have:
    \begin{equation*}
      \min_{\bfbeta\in\mathbb{R}^{p}} \max_{\bfgamma' \in \uball[*]} 
      J_\lambda(\bfbeta,\bfgamma') 
      \geq
      \frac{\eta}{\norm[*]{\bfgamma}} 
      J_\lambda\left(\bfbeta^\star\left(\bfgamma\right), \bfgamma \right) -
      \frac{\lambda\eta(\norm[*]{\bfgamma}-\eta)}{\norm[*]{\bfgamma}^2}\norm{\bfgamma}^2        
      \enspace,
    \end{equation*}
    where 
    \begin{equation*}
      J_\lambda(\bfbeta,\bfgamma) = \norm{\bfX \bfbeta - \bfy}^2 + 
        \lambda \norm{\bfbeta - \bfgamma}^2
      \enspace \text{and} \enspace
      \bfbeta^\star(\bfgamma) = \argmin_{\bfbeta\in\mathbb{R}^{p}} J_\lambda(\bfbeta,\bfgamma)
      \enspace.
    \end{equation*}
    See proof in  \ref{sec:proof:prop:monitoring}.
  \end{proposition}

  This proposition can be used to compute an optimality gap at Step 3 of
  Algorithm \ref{algo:active_set}, by picking a $\bfgamma$-value such that the
  current worst-case gradient $\bfg$ is null (the current $\bfbeta$-value then
  being the optimal $\bfbeta^\star(\bfgamma)$).
  Note that more precise upper bounds could be computed relying on significant extra computation.
  The generic optimality gap computed from Proposition \ref{prop:monitoring} differs
  from the Fenchel duality gap \citep[see][]{2012_FML_Bach}. 
  For the elastic net expressed in \eqref{eq:elastic-net}, Fenchel inequality
  \citep[see details in][]{Mairal10} yields the following optimality gap:
   \begin{align*}
      \min_{\bfbeta\in\mathbb{R}^{p}} \max_{\bfgamma' \in \uball[*]^\eta} 
      J_\lambda(\bfbeta,\bfgamma') 
      \geq &
      J_\lambda\left(\bfbeta^\star\left(\bfgamma\right), \bfgamma \right) -
      \frac{\eta^{2}}{\norm[*]{\bfgamma}^{2}}
      \left(
        \norm{\bfX \bfbeta^\star\left(\bfgamma\right) - \bfy}^2 + 
        \lambda \norm{\bfbeta^\star\left(\bfgamma\right)}^{2}
      \right) \\
      & - \frac{2\eta}{\norm[*]{\bfgamma}}
      \left(
         \bfX \bfbeta^\star\left(\bfgamma\right) - \bfy
     \right)^\intercal \bfy
     \enspace.
   \end{align*}
  \begin{figure}
    \centering 
    \xylabellarge{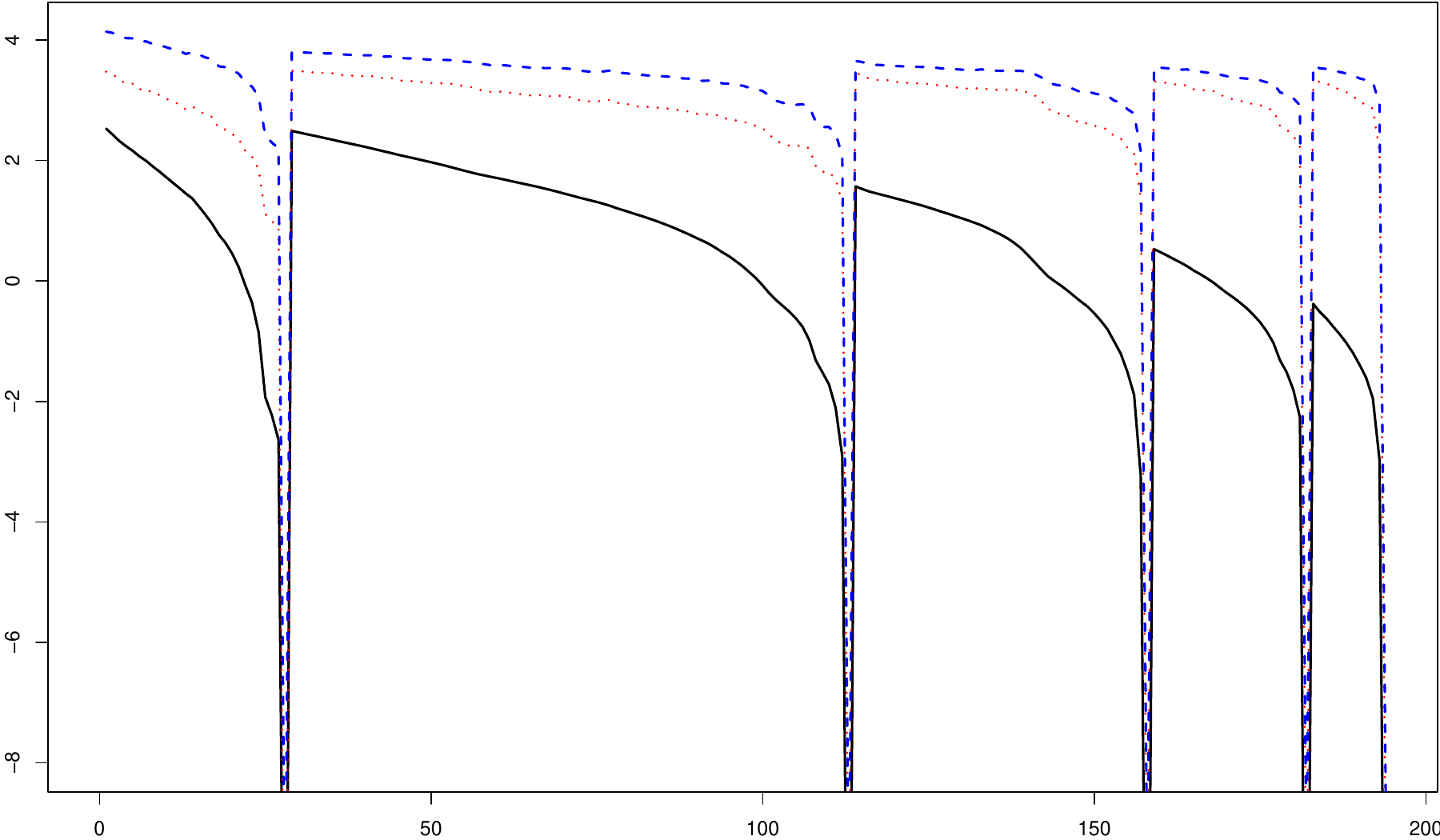}{\# of iterations}{Optimality gap (log scale)}{$n=50$, $p=200$}
    \caption{Monitoring convergence: true optimality gap (solid black) versus
    our generic upper bound (dashed blue) and Fenchel's duality gap for elastic
    net (dotted red) computed at each iteration of
    Algorithm~\ref{algo:active_set}.}
    \label{fig:monitoring}
  \end{figure}
  The two optimality gaps are empirically compared in 
  Figure~\ref{fig:monitoring} for the elastic net, along a short regularization 
  path with five values of the $\ell_{1}$-penalization parameter.  
  We see that the two options can be used to assess convergence, though
  Fenchel's duality gap is more accurate for the rougher solutions. 
  Note however that both upper bounds are fairly coarse until a very accurate
  solution is reached, which makes them both unsuitable for deriving loose stopping 
  criteria.
  Proposition~\ref{prop:monitoring} is thus of limited scope, but it illustrates
  that, besides its algorithmic consequences, our original view of sparse
  penalties opens new ways for analysis.
  As a final note on this topic, we provide a slightly tighter inequality for
  computing the optimality gap in~\ref{sec:proof:prop:monitoring}, and
  we conjecture that it could be further tightened (see in particular the
  derivation of inequality \eqref{eq:crude_inequality}).

\section{Numerical Experiments}\label{sec:experiments}

This section compares the performances of our algorithm to its state-of-the-art
competitors from an optimization viewpoint.  Efficiency may then be assessed by
accuracy and speed:  accuracy is the difference between the optimum of the
objective function and its value at the solution returned by the algorithm;
speed is the computing time required for returning this solution.
Obviously, the timing of two algorithms/packages has to be compared at similar
precision requirements, which are rather crude in statistical learning, far from
machine precision \citep{Bottou08}.

We start by showing an experiment on a real dataset illustrating that optimization is an actual issue from the data analysis viewpoint. We then provide an in-depth analysis on simulated data, thereby covering a wide range of well-controlled situations, where  our conclusions are derived from numerous simulations.
Finally, the available ground truths will also be useful to show that optimization issues during training have notable effects in terms of prediction accuracy and support recovery.

\subsection{Introductory Example: Quantitative Trait Loci and Association Mapping}

Many quantitative traits in plants and animals are heritable.  When
considering Mendelian traits, the Quantitative Trait Loci (QTL) are sections of
DNA that explain the trait variability.
As traits are typically controlled by several QTL, the association 
between traits and QTL involves several limited effects that are difficult to detect individually.

A possible method for mapping genotype to trait consists in regressing
the trait (that is, phenotype) of interest  against the QTL (that is, genotype), using penalized regression.
In this introductory example, we consider the maize association
mapping panel described in \citep{RincentEtAl2014},where 269
individuals were genotyped with a 50k single-nucleotide polymorphisms (SNPs) array. 
After classical data
cleaning, 261 individuals with 29849 markers (that is, SNPs) were kept.

We focus here on the tasseling time of maize, whose heritability is explained by many genes.
We used two different implementations of Lasso for detecting the relevant markers: \mytexttt{glmnet} \citep[Generalized Linear
Models regularized by Lasso and elastic-NET,][]{2009_JSS_Friedman} and
our own implementation, 
publicly available through the
\mytexttt{R} package
\mytexttt{quadrupen}%
\footnote{The \mytexttt{quadrupen} package is available on the CRAN (Comprehensive R Archive Network
  {\url{https://cran.r-project.org/web/packages/quadrupen/}}}.

The two implementations were run using their default parameters, with the 
same penalty strength $\lambda$, selected by leave-one-out cross-validation with 
\mytexttt{glmnet}.
The latter selected 169 markers among the 29849 SNPs, whereas \mytexttt{quadrupen} was more stringent, selecting only a subset of 156 markers. 
This notable difference is unexpected considering that the two implementations attempt to solve the very same convex optimization problem.

\begin{figure}
  \centering
  \begin{tabular}{lc}
  \rotatebox{90}{\small \hspace{2.25cm} count}
    & \includegraphics[width=.6\textwidth]{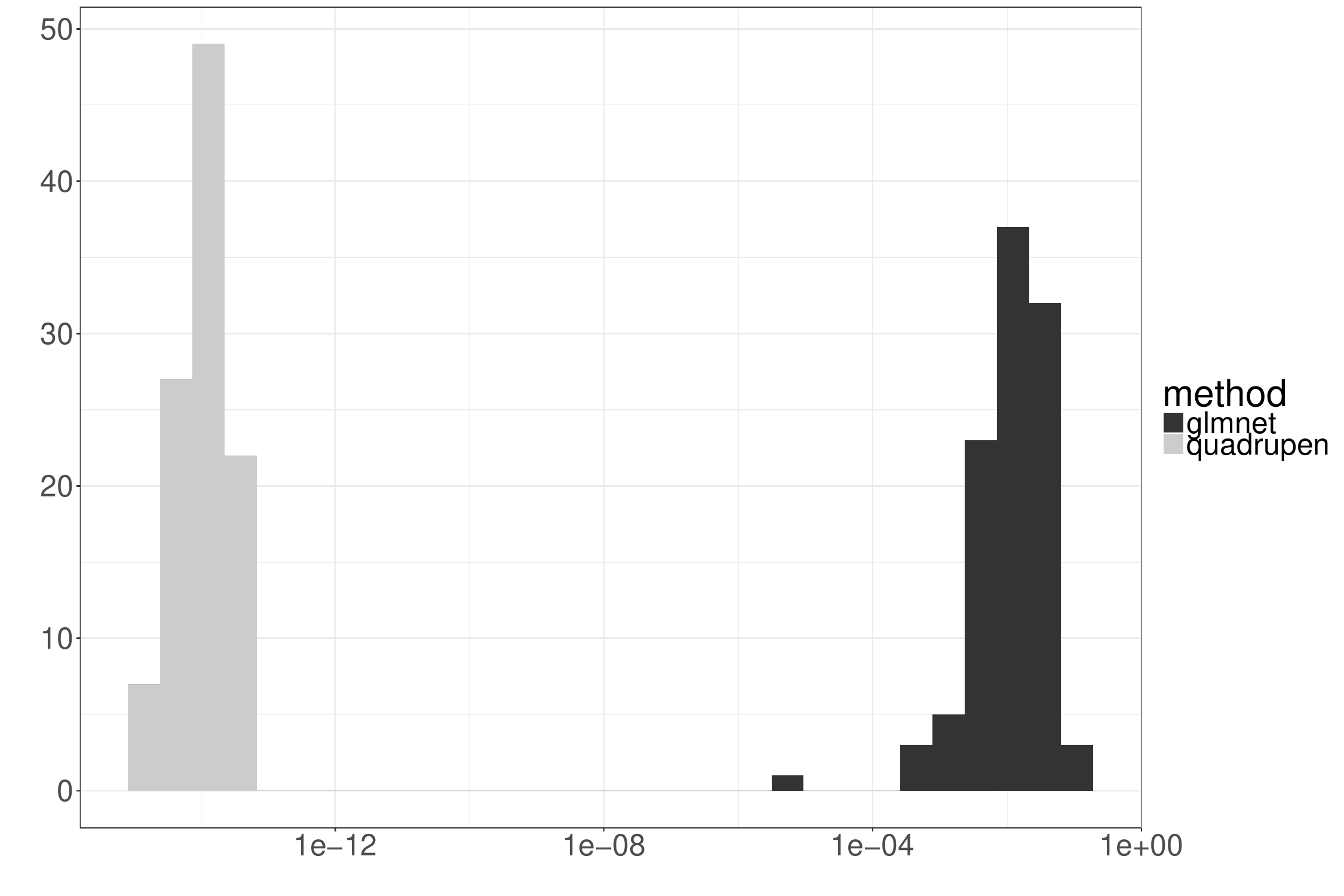} \\
    & \small $\left| \mathbf{x}_j^\intercal\left(\mathbf{y}  -\mathbf{X}\hatbbeta\right) \right| -  \lambda$ \hspace{1cm} (log scale) \\
  \end{tabular}
  \caption{Histogram of the derivatives w.r.t. the non-zero coefficients of the parameter vectors returned by \mytexttt{glmnet} and \mytexttt{quadrupen}}.
  \label{fig:gradient}
\end{figure}

In the following, we show how the speed and precision of \mytexttt{glmnet} are affected by the threshold controlling the stopping condition. 
Regarding precision, a first hint is provided in 
Figure \ref{fig:gradient}, which displays the absolute value of the derivatives of the objective function with respect to the non-zero coefficients of the parameter vectors returned by \mytexttt{glmnet} and \mytexttt{quadrupen}, respectively. 
The departure from zero for \mytexttt{glmnet} illustrates its relative imprecision, which is responsible for the difference in the set of selected markers.

\subsection{Synthetic data}

We compare the performance of the proposed quadratic solver with
representatives of the most successful competing algorithms.
First, we use our own implementations of all competitors, so as to provide
comparisons without implementation biases (language, library, etc.).
We use \mytexttt{R} with most of the matrix calculus done in \mytexttt{C++} using
the \mytexttt{RcppArmadillo} package \citep{2011_JSS_rcpp,armadillo} that relies
on \mytexttt{BLAS/LAPACK} libraries for linear algebra operations.
Second, we compare our code to the leading standalone packages that are
available today, so as to provide comparisons avoiding a possible competence
bias.

We use simulated data to obtain representative average results. 
Their generation covers the typical attributes of the real data encountered in
post genomic and signal processing.
In these domains, the main optimization difficulties result from
ill-conditioning, which is either due to the high correlation between
predictors, or to underdetermination when the number of variables
exceeds the sample 
\iflong 
  size (also known as the high-dimensional or the ``large $p$ small $n$'' 
  setup).
\fi
For the optimization algorithms based on active set strategies, bad
conditioning is somehow alleviated when the objective function has a regular
behavior when restricted to the subspace containing the solution.  All other
things being equal, this local conditioning is thus governed by the
sparsity of the unknown true parameter (which affects the sparsity of
the solution), which also heavily impacts the running times of most
optimization algorithms available today.

\subsubsection{Data Generation}

The above-mentioned characteristics are explored without difficulty in the
framework of linear regression. We generate samples of size $n$ from the model
\begin{equation*}
  \mathbf{y} = \mathbf{X} \bfbeta^\star + \varepsilon, \qquad
  \varepsilon \sim\mathcal{N}(\mathbf{0},\sigma^2\mathbf{I})
  \enspace,
\end{equation*}
with $\sigma$ chosen so as to reach a rather strong  coefficient of
determination ($R^2\approx 0.8$).  The design matrix $\mathbf{X}$ is drawn from
a multivariate normal distribution in $\Rset^p$, and the conditioning of
$\mathbf{X}^\intercal\mathbf{X}$ is ruled by the correlation between variables. 
We use the same correlation coefficient $\rho$ for all pairs of variables.
The sparsity of the true regression coefficients is controlled by a parameter
$s$, with
\begin{equation*}
  \bfbeta^\star = \big(\underbrace{2,\dots,2}_{s/2} , \underbrace{-2,\dots,-2}_{s/2} , \underbrace{0,\dots,0}_{p-s} \big)
  \enspace.
\end{equation*}
Finally, the ratio $p/n$
quantifies the well/ill-posedness of the problem.

 
\subsubsection{Comparing Optimization Strategies} 

We compare here the performance of three state-of-the-art optimization
strategies implemented in our own computational framework:
accelerated proximal method \citep[see, e.g.,][]{2009_SIAM_Beck},
coordinate descent \citep[popularized by][]{2007_AAS_Friedman},
and our algorithm, that will respectively be named hereafter \mytexttt{proximal},
\mytexttt{coordinate} and \mytexttt{quadratic}.
Our implementations estimate the solution to the elastic net problem
\begin{equation}
  \label{eq:enet}
    J_{\lambda_1,
    \lambda_2}^{\text{enet}}(\boldsymbol\beta)= \frac{1}{2}
  \norm{\mathbf{X}\boldsymbol\beta - \mathbf{y}}^2 +
  \lambda_1 \norm[1]{\boldsymbol\beta} + \frac{\lambda_2}{2} \norm{\boldsymbol\beta}^2
  \enspace,
\end{equation}
which is strictly convex when $\lambda_2>0$ and thus admits a unique
solution even if $n<p$.

The three implementations are embedded in the same active set routine, which
approximately solves the optimization problem with respect to a limited number
of variables as in Algorithm~\ref{algo:active_set}.
They only differ regarding the inner optimization problem with respect to the
current active variables, which is performed by an
accelerated proximal gradient method for \mytexttt{proximal}, by coordinate descent
for  \mytexttt{coordinate}, and by the resolution of the worst-case quadratic
problem for \mytexttt{quadratic}.
We followed the practical recommendations of \citet{2012_FML_Bach} for
accelerating the proximal and coordinate descent implementations, and we used
the same halting condition for the three implementations, based on the
approximate satisfaction of the first-order optimality conditions:
\begin{equation}
  \label{eq:tau_conv}
  \max_{j\{\in 1\dots p\}} \left| \mathbf{x}_j^\intercal\left(\mathbf{y}
      -\mathbf{X}\hatbbeta                 
    \right) + \lambda_2\hatbbeta \right| <
  \lambda_1 + \tau,
\end{equation}
where the threshold $\tau$ was fixed to $\tau=10^{-2}$ in our simulations.~\footnote{%
  The rather loose threshold is favorable to \mytexttt{coordinate} and
  \mytexttt{proximal}, which reach the threshold, while \mytexttt{quadratic}
  ends up with a much smaller value, due to the exact resolution, up to
  machine precision, of the inner quadratic problem.
}
Finally, the active set algorithm is itself wrapped in a warm-start routine,
where the approximate solution to $J^{\text{enet}}_{\lambda_1,\lambda_2}$ is
used as the starting point for the resolution of
$J^{\text{enet}}_{\lambda_1',\lambda_2}$ for $\lambda_1' < \lambda_1$.

Our benchmark considers small-scale problems, with size $p=100$, and the nine
situations stemming from the choice of three following parameters:
\iflong
\begin{itemize}
\item low,  medium  and  high correlation between predictors ($\rho \in \{0.1, 0.4, 0.8\}$),
\item low, medium and  high-dimensional setting ($p/n \in \{2, 1,
  0.5\}$),
\item low, medium and high levels of sparsity ($s/p\in\{0.6 ,0.3,0.1\}$).
\end{itemize}
\fi
Each solver computes the elastic net for the tuning parameters $\lambda_1$ and
$\lambda_2$ on a 2D-grid of $50 \times 50$ values, and their running 
times have been averaged over 100 runs.

All results are qualitatively similar regarding the dimension and sparsity
settings.
Figure~\ref{fig:timing_all} displays the high-dimensional case ($p=2n$) with
a medium level of sparsity ($s=30$) for the three levels of correlation.
\iflong
  \begin{figure}
    \centering
    \setlength{\unitlength}{0.575\linewidth}%
    \begin{picture}(1.5,2.2)%
      \put(0,0){\includegraphics[angle=90,width=1.45\unitlength]{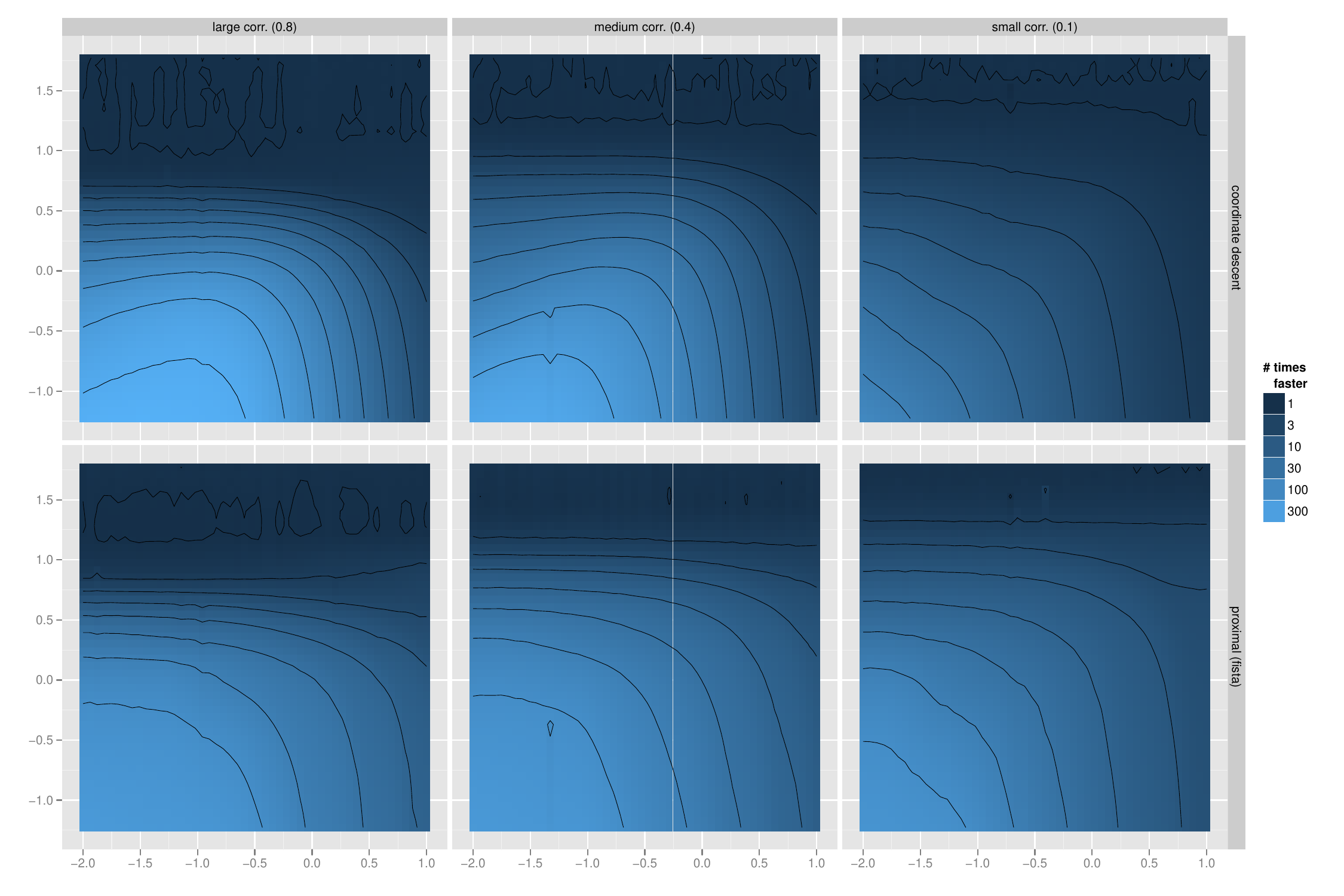}}
      \put(1.45,1.65){\rotatebox{90.0}{\makebox[0cm]{$\log_{10}{\lambda_1}$}}}
      \put(1.45,1.05){\rotatebox{90.0}{\makebox[0cm]{$\log_{10}{\lambda_1}$}}}
      \put(1.45,0.45){\rotatebox{90.0}{\makebox[0cm]{$\log_{10}{\lambda_1}$}}}
      \put(0.4,0){\makebox[0cm]{$\log_{10}{\lambda_2}$}}
      \put(1.025,0){\makebox[0cm]{$\log_{10}{\lambda_2}$}}
    \end{picture} 
     \caption{Log-ratio of computation times for \mytexttt{coordinate} (left), and
     \mytexttt{proximal} (right), versus \mytexttt{quadratic}, for
     $(p,n)=(100,50),\, s=30$ and correlation $\rho \in \{0.1, 0.4, 0.8\}$
     (top, middle and bottom respectively).}
    \label{fig:timing_all}
  \end{figure}
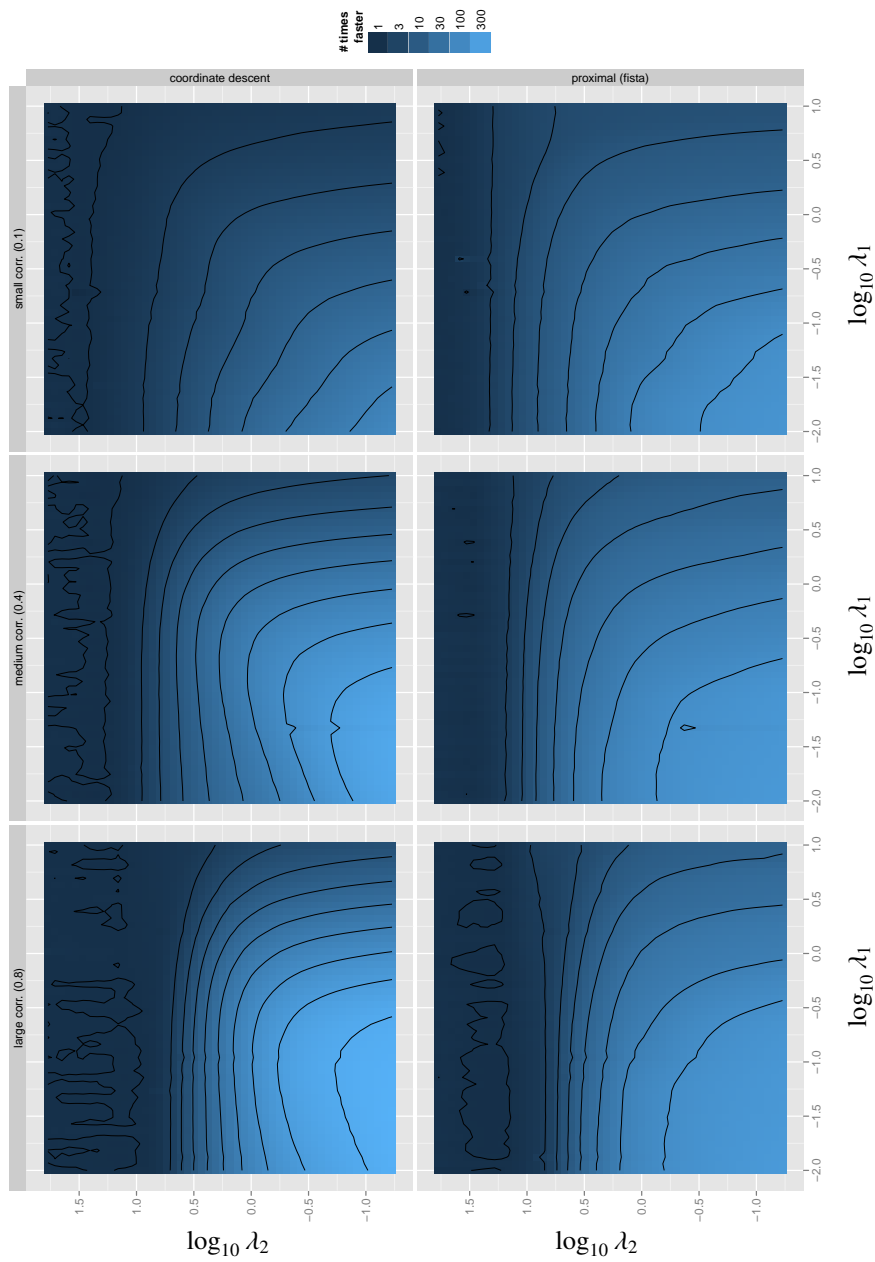
\fi
Each map represents the log-ratio between the timing of either
\mytexttt{coordinate} or \mytexttt{proximal} versus \mytexttt{quadratic},
according to $(\lambda_1, \lambda_2)$ for a given correlation level.  
Dark regions with a value of 1 indicate identical running times while lighter
regions with a value of 10 indicate that \mytexttt{quadratic} is 10 times faster.
Figure~\ref{fig:timing_all} illustrates that \mytexttt{quadratic} outperforms both
\mytexttt{coordinate} and \mytexttt{proximal}, by running much faster in most
cases, even reaching 300-fold speed increases.  
The largest gains are observed for small $(\lambda_1,\lambda_2)$ penalty
parameters for which the problem is ill-conditioned, including many active
variables, resulting in a huge
slowdown of the first-order methods \mytexttt{coordinate} and
\mytexttt{proximal}.
As the penalty parameters increase, smaller gains are observed, especially when
$\lambda_2$, attached to the quadratic penalty, reaches high values for which all
problems are well-conditioned, and where the elastic net is leaning towards
univariate soft thresholding, in which case all algorithms behave similarly.

\subsubsection{Comparing Stand-Alone Implementations}

We  now  proceed  to the  evaluation  of  our  code with  three  other
stand-alone programs publicly  available as \mytexttt{R} packages.  We
chose     three    leading    state-of-the-art     packages,    namely
\mytexttt{glmnet}  \citep[Generalized  Linear  Models  regularized  by
Lasso    and    elastic-NET,][]{2009_JSS_Friedman},    \mytexttt{lars}
\citep[Least      Angle     Regression,     lasso      and     forward
Stagewise,][]{2004_AS_Efron}    and   \mytexttt{SPAMS}   \citep[SPArse
Modeling     Software,][]{2012_FML_Bach},     with     two     options
\mytexttt{SPAMS-FISTA},  which   implements  an  accelerated  proximal
method,   and  \mytexttt{SPAMS-LARS}   which   is  a   \mytexttt{lars}
substitute.   Note that  \mytexttt{glmnet} does  most of  its internal
computations  in \mytexttt{Fortran}, \mytexttt{lars}  in \mytexttt{R},
and  \mytexttt{SPAMS} in \mytexttt{C++}.   


We benchmark these packages by computing  regularization paths for the
Lasso\footnote{%
  We benchmark the packages on a Lasso problem since the parametrization of the
  elastic net problem differs among packages, hindering fair comparisons.},
that is, the elastic net Problem~\eqref{eq:enet} with $\lambda_2=0$.   
The inaccuracy of the solutions produced is measured by the gap in the objective
function compared to a reference solution, considered as being the true optimum.  
\iflong
  We use the \mytexttt{lars} solution as a reference, since it solves the Lasso problem
  up to the machine precision, relying on the \mytexttt{LAPACK/BLAS} routines. 
  Furthermore, \mytexttt{lars} provides the solution path for the Lasso,
  that is, the set of solutions computed for each penalty parameter value
  for which variable activation or deletion occurs, from the empty model to the
  least-mean squares model.
  This set of reference penalty parameters is used here to define a sensible reproducible choice.

  In high dimensional setups, the computational cost of returning the solutions
  for the largest models may be overwhelming compared to the one necessary
  for exploring the interesting part of the regularization path
  \citep{2011_JSS_Simon,2009_JSS_Friedman}.  This is mostly due to numerical
  instability problems that may be encountered in these extreme settings, where
  the Lasso solution is overfitting as it approaches the set of solutions to the
  underdetermined least squares problem.  We avoid a
  comparison mostly relying on these spurious cases by restricting the set
  of reference penalty parameters to the first $\min(n,p)$ steps of
  \mytexttt{lars} \citep[similar settings are used by][]{2009_JSS_Friedman}.

  Henceforth, the  distance $\mathrm{D}$  of a given  \mytexttt{method} to
  the optimum is evaluated on  the whole set of penalties $\Lambda$ used
  along the path, by
  \begin{equation*}
    \mathrm{D}(\mytexttt{method}) = \left( \frac{1}{|\Lambda|}\sum_{\lambda\in\Lambda}
      \left(J_{\lambda}^{\text{lasso}}\left(\hatbbeta_\lambda^{\mytexttt{lars}}\right)
        -J_{\lambda}^{\text{lasso}}\left(\hatbbeta_\lambda^{\mytexttt{method}}\right)\right)^2
       \right)^{1/2} 
    \enspace,
  \end{equation*}
  where $J_{\lambda}^{\text{lasso}}(\bfbeta) = J_{\lambda,0}^{\text{enet}}(\bfbeta)$ 
  is the objective function of the Lasso evaluated at $\bfbeta$, and
  $\hatbbeta_\lambda^{\mytexttt{method}}$ is the estimated optimal solution
  provided by the \mytexttt{method} package currently tested.
\fi

The data  sets are  generated according to  the linear  model described
above, in three different high-dimensional settings and small to medium number 
of variables: 
$(p,n)=(100,40)$, $(p,n)=(1\,000,200)$ and $(p,n)=(10\,000,400)$.  
The sparsity of the true underlying $\boldsymbol\beta^\star$ is governed by $s =
0.25\min(n,p)$, and the correlation between predictors is set by $\rho\in\{0.1, 0.4,  0.8\}$. For  each  value of
$\rho$, we averaged the timings over $50$ simulations, ensuring that each package
computes the solutions at identical $\lambda$ values, as defined above.

\iflong

  We pool together the runtimes obtained for the three levels of correlation for
  \mytexttt{quadrupen}, \mytexttt{SPAMS-LARS} and \mytexttt{lars}, which are not
  sensible to the correlation between features. 
  In each plot of Figure~\ref{fig:timing_glmnet}, each of these methods is thus 
  represented by a single point marking the average precision and the average distance to the
  optimum over the 150 runs (50 runs for each $\rho \in \{0.1, 0.4, 0.8\}$).
  Note that for \mytexttt{lars} only the abscissa is meaningful since
  $\mathrm{D}(\mytexttt{lars})$ is zero by definition.
  Besides, \mytexttt{quadrupen}, which solves each quadratic problem up to the
  machine precision, tends to be within this precision of the \mytexttt{lars}
  solution.
  The \mytexttt{SPAMS-LARS} is also very precise, up to $10^{-6}$,
  which is the typical precision of the approximate resolution of linear systems.
  It is the fastest alternative for solving the Lasso when the problem is
  high-dimensional with a large number of variables (Figure
  \ref{fig:timing_glmnet}, bottom-left).

  \begin{figure}
    \centering
    \begin{tabular}{cc}
      \xylabelsquare{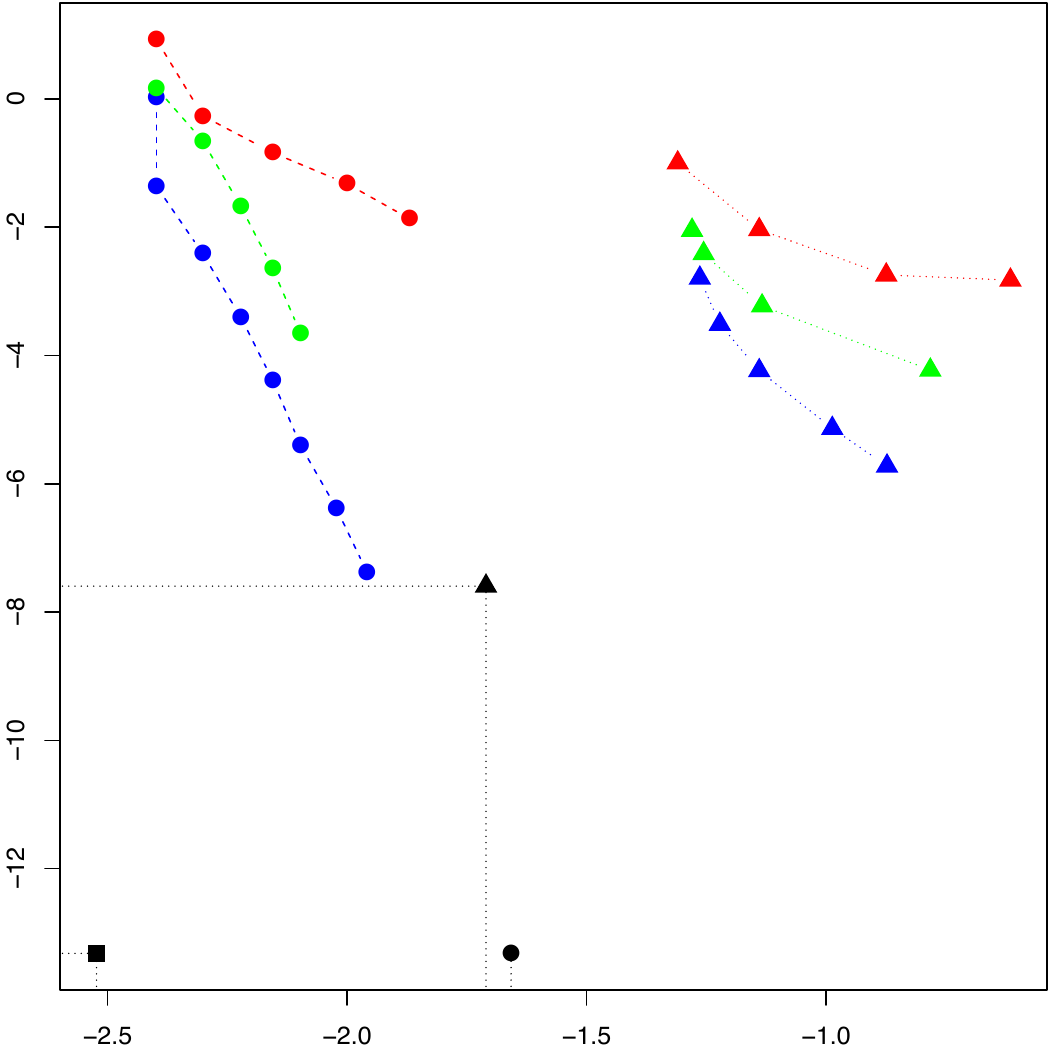}{CPU time (in seconds, $\log_{10}$)}
                    {$\mathrm{D}(\mytexttt{method})$ ($\log_{10}$)}
                    {$n=100$, $p=40$}%
      & \xylabelsquare{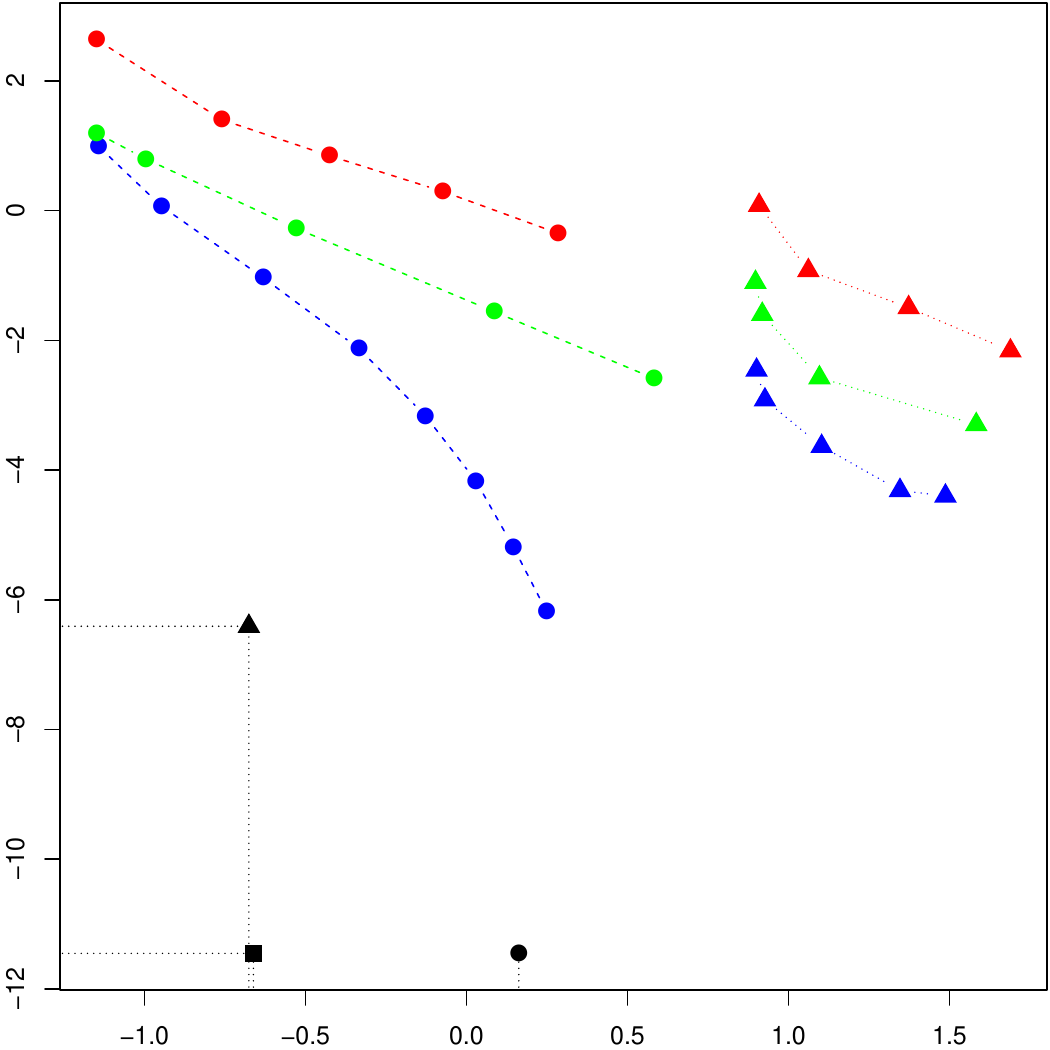}{CPU time (in seconds, $\log_{10}$)}
                    {$\mathrm{D}(\mytexttt{method})$ ($\log_{10}$)}
                    {$n=200$, $p=1\,000$}%
      \\[4ex] %
      \xylabelsquare{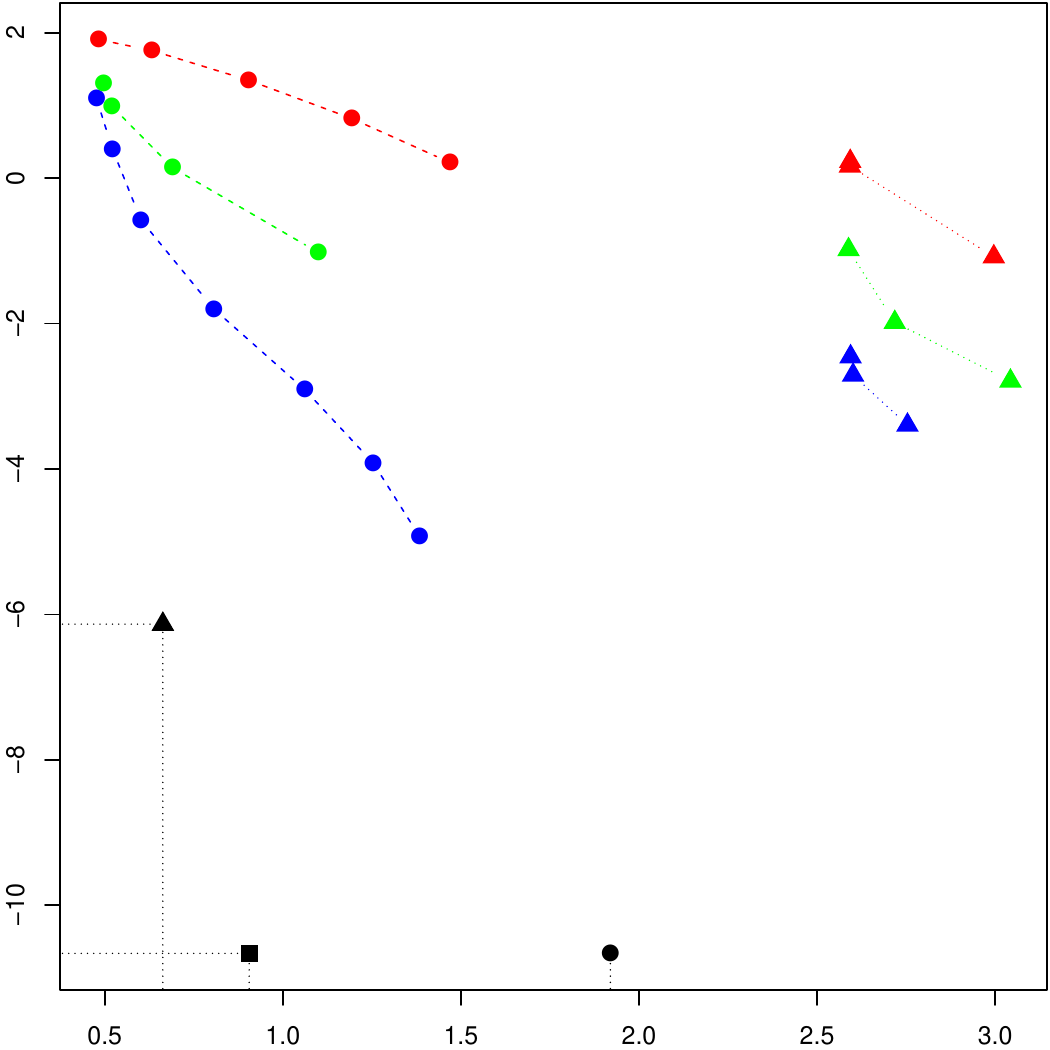}{CPU time (in seconds, $\log_{10}$)}
                    {$\mathrm{D}(\mytexttt{method})$ ($\log_{10}$)}
                    {$n=400$, $p=10\,000$}%
      & \xylabelsquare{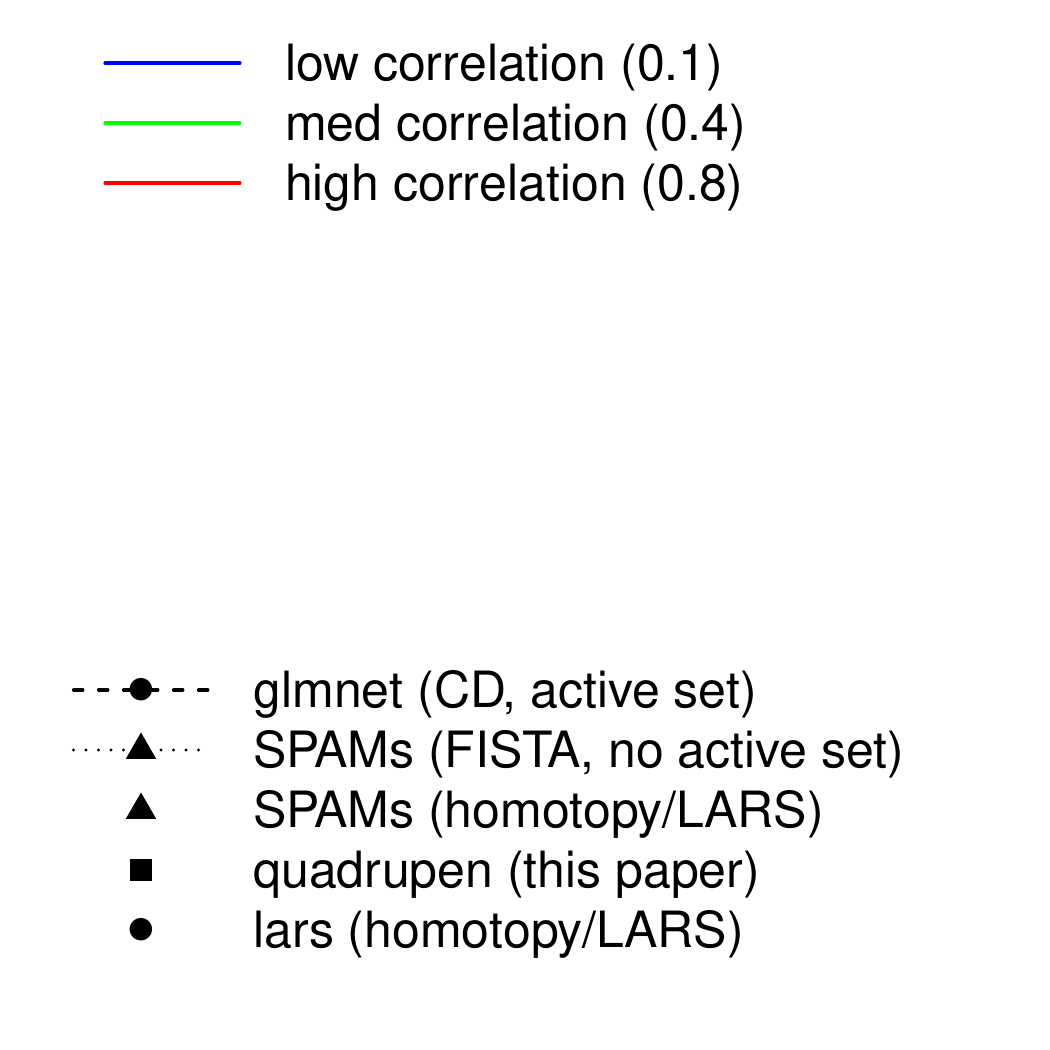}{}{}{} \\
    \end{tabular}
    \caption{Distance  to optimum  versus CPU  time for  three different
     high-dimensional settings: $(p,n)=(100,40)$ (top left), $(p,n)=(1\,000,200)$
     (top right) and $p=(10\,000,400)$ (bottom left).  }
    \label{fig:timing_glmnet}
  \end{figure}

  In contrast, the (precision,timing)-values of \mytexttt{glmnet} and
  \mytexttt{SPAMS-FISTA} are highly affected by the threshold
  parameters\footnote{%
    In \mytexttt{glmnet}, convergence is monitored by the stability of the
    objective function, measured between two optimization steps, and
    optimization is halted when changes fall below the specified threshold
    (scaled by the null deviance).
    In \mytexttt{SPAMS-FISTA}, the stopping condition of the algorithm is based
    on the relative change of parameters between two iterations.}
  that control their stopping conditions.
  The computational burden to reach a given precision is also affected by the
  level of correlation, as illustrated in Figure~\ref{fig:timing_glmnet}.
  Obviously, a precise solution is difficult to reach with first-order descent
  algorithms in a high correlation setup, which corresponds to an
  ill-conditioned linear system.
  It may be surprising to observe that \mytexttt{SPAMS-FISTA} is about ten time
  slower than \mytexttt{glmnet}, as proximal and coordinate descent
  methods were experimentally shown to be roughly equivalent in our preceding
  analysis and by \citet{2012_FML_Bach}.
  However, these two comparisons were carried out with the same active set 
  strategy \citep[that is, \emph{with} active set for ours and \emph{without} 
  active set for][]{2012_FML_Bach}.
  We believe that this difference in the handling of active variables explains
  the relative bad performance of \mytexttt{SPAMS-FISTA}, which optimizes all
  variables along the regularization path, while  \mytexttt{glmnet} uses a 
  greedy active set strategy. 

 Overall, our implementation is highly competitive, that is, very accurate, at
 the \mytexttt{lars} level, and much faster.  The speed improvements of
 \mytexttt{glmnet} are only observed for very rough approximate solutions and
 \mytexttt{SPAMS-FISTA} is dominated by \mytexttt{glmnet}. 
 Our experiments, in the framework of active set methods, agree with the results of
 \citet{2012_FML_Bach}: indeed, they observed that first-order methods are
 competitive with second-order ones only for low correlation levels and
 small penalties (which entails a large number of active variables).
 Conversely, our results may appear to contradict some of the experimental
 findings of \citet{2009_JSS_Friedman}: first, we observe that \mytexttt{glmnet}
 is quite sensitive to correlations, and second, the optimized second-order
 methods are competitive with \mytexttt{glmnet}.
 These differences in conclusions arise from the differences in experimental
 protocols: while we compare running times at a given accuracy, they are compared
 at a given threshold on the stopping criterion by \citet{2009_JSS_Friedman}.
 Regarding the influence of correlations, the stability-based criterion can be
 fooled due to the tiny step size that typically occurs for ill-conditioned
 problems, leading to a sizable early stopping.
 Regarding the second point, even though the \mytexttt{R} implementation of
 \mytexttt{lars} may indeed be slow compared to \mytexttt{glmnet}, considerable
 improvements can be obtained using optimized second-order methods such as 
 \mytexttt{quadrupen} as soon as a sensible accuracy is required, especially
 when correlation increases.
 
 Finally, among the accurate solvers, \mytexttt{SPAMS-LARS} is insignificantly
 less accurate than \mytexttt{quadrupen} or \mytexttt{lars} in a statistical
 context.  It is always faster than \mytexttt{lars} and slightly faster than
 \mytexttt{quadrupen} for the largest problem sizes (Figure
 \ref{fig:timing_glmnet}, bottom-left) and much slower for the smallest problem
 (Figure \ref{fig:timing_glmnet}, top-left).

\fi

\subsubsection{Link between accuracy of solutions and prediction performances}
\label{sec:fromatop}

When the ``irrepresentable condition'' \citep{2006_JMLR_Zhao} holds, the
Lasso should  select the true model consistently.   However, even when
this  rather  restrictive  condition  is  fulfilled,  perfect  support
recovery  obviously requires numerical  accuracy:
rough estimates may speed up the procedure, but whatever optimization strategy
is used, stopping an algorithm is likely to prevent either the removal of all
irrelevant coefficients or the insertion of all relevant ones.  The support of 
the solution may then be far from the optimal one. 

\begin{figure}[htbp]
  \centering
  \begin{tabular}{@{}l@{}c@{}} 
    \rotatebox{90.0}{\makebox[.6\textwidth]{\hspace{.3\textwidth} Support Error 
    Rate \hspace{.35\textwidth} MSE}}
    &
    \includegraphics[width=.95\textwidth]{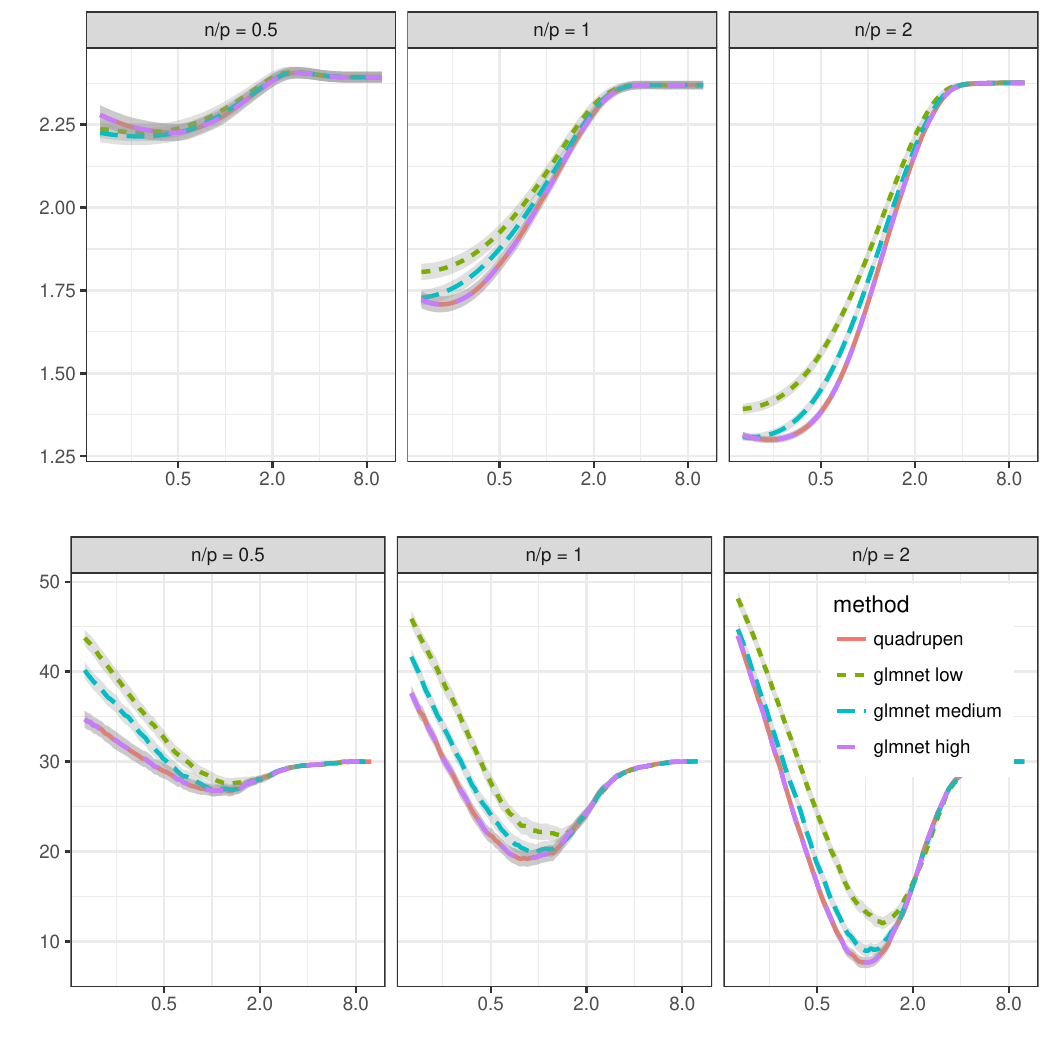}
  \end{tabular}
  \caption{Test performances according to the penalty parameter for the
    Lasso estimates returned by \mytexttt{quadrupen} and \mytexttt{glmnet} at various level
    of accuracy. Three high-dimensional setups are illustrated: from left to 
    right $n/p=1/2$, $n/p=1$ and $n/p=2$;
    top: mean squared test error;
    bottom: support error rate.\label{fig:accuracy}}
\end{figure}

We  advocate here  that our  quadratic solver  is very  competitive in
computation time when support recovery matters, that is, when high level  of accuracy is needed, in small (few
hundreds of variables) and  medium sized problems (few thousands).  As
an  illustration,  we  
generate 100 data sets under the linear model described above, with a rather
strong coefficient of determination ($R^2
\approx 0.8$ on  average), a rather high level  of correlation between
predictors ($\rho=0.8$) and a medium level of sparsity ($s/p = 30\%$).
The number of  variable is kept low ($p=100$) and  the difficulty of the
estimation problem  is tuned  by the  $n/p$  ratio. For  each data  set, we  also
generate a  test set sufficiently  large (say, $10n$) to  evaluate the
quality of the prediction without  depending on any sampling fluctuation.  
We compare the Lasso solutions computed by \mytexttt{quadrupen} to the ones returned by
\mytexttt{glmnet} with
various   level   of    accuracy\footnote{This   is   done   via   the
  \mytexttt{thresh}  argument of  the  \mytexttt{glmnet} procedure,  whose
  default  value  is \mytexttt{1e-7}.   In  our  experiments,
  \mytexttt{low}, \mytexttt{med}  and \mytexttt{high}  level of  accuracy for
  \mytexttt{glmnet} respectively correspond to \mytexttt{thresh} set to
  \mytexttt{1e-1}, \mytexttt{1e-4}, and \mytexttt{1e-9}.}.  
Figure \ref{fig:accuracy} reports performances, as measured by the mean squared
test error and the support error rate.

As expected, the curves show that selecting variables and searching for
the best prediction are two different problems.  The selection problem
(bottom  of Figure \ref{fig:accuracy}) always  requires a sparser  model than the
prediction  problem.  But  despite this  obvious difference,  the more
accurate the solution returned by the algorithm, the better  the performances for any levels of
penalty and for  both performance measures. 

\begin{table}[htbp]

  \begin{tabular}{@{}l|cccc@{}}
    methods & \mytexttt{quadrupen} & \mytexttt{glmnet\,low} & \mytexttt{glmnet\,med} & \mytexttt{glmnet\,high} \\
    \hline
    timing (ms) & 8 & 7 & 8 & 64 \\
    accuracy (dist.  to opt.)  & $5.9\times 10^{-14}$ & $7.2 \times 10^{0}$ & $6.04 \times 10^{0}$ & $1.47 \times 10^{-2}$\\
  \end{tabular}
  \caption{Median timings and solution accuracies
  }
  \label{tab:accuracy}
\end{table} 

Now focusing on \mytexttt{glmnet} performances, the better the accuracy,
the  smaller the  MSE  and the  support  error rate.  But  the better  the
accuracy,  the  slower  the  algorithm  becomes.   Using  the  default
settings allows to have a result very close to our quadratic solver, and the
performance differences become negligible between our approach and \mytexttt{glmnet}
running with high precision.
However, Table \ref{tab:accuracy} illustrates that high accuracy is achieved at 
a high computational cost: to be at par with \mytexttt{quadrupen} with regards to
test performances, \mytexttt{glmnet} is about ten times slower than our solver.

\section{Discussion}


This paper presents a new viewpoint on sparsity-inducing penalties
where the dual norms associated with these penalties play a central role. 
Technically, our formulation is a simple dual form of the original problem. 
However, we do not follow a very general principle such as Fenchel duality: 
we specifically tailor our formulation to optimization problems involving a 
sparsity-inducing norm.
The dual variables define a series of linear or quadratic penalties whose 
sublevel sets define the feasible set of the original problem through
intersection.

This  viewpoint  enables  to  encompass in the same framework  several
well-known penalties. In particular, we detailed how the solutions to the Lasso and the
group-Lasso (with the $\ell_{\infty,1}$  mixed norm), possibly applied
together with an  $\ell_2$ ridge penalty (leading to what  is known as
the elastic net for the Lasso) can be derived. 

We derived a  general-purpose algorithm that computes the solution to
the penalized regression problem,  and proposed a new lower
bound  on the  minimum  of  the objective  function to
assess convergence.  The proposed  algorithm solves a series of
quadratic  problems on a working set defined  by the dual  variables.   
It has  been
thoroughly tested  and compared with  state-of-the-art implementations
for the elastic net and the Lasso, and prevails over its competitors for 
the problems tested, involving  up to a few thousands of variables.

From a  practical viewpoint, an  important feature of our  approach is
that it solves the original problem  up to machine precision: we shown
that when  variable selection  is involved,  optimization with  a high
level of precision is mandatory to recover the true model.
\\

Regarding  future  development,  the   algorithm  can  be  adapted  to
non-quadratic loss  functions for  addressing other  learning problems
such  as  classification,  but  this  generalization,  which  requires
solving non-quadratic  problems, may not  be as efficient  compared to
the  existing alternatives.   We are  now examining  how to  address a
wider range of penalties by extending the framework in two directions:
first,  to accommodate  additional general  $\ell_2$ penalties  in the
form of  arbitrary symmetric  positive semidefinite matrix  instead of
the simple ridge, in particular to provide an efficient implementation
of the  structured elastic  net \citep{2010_AOS_Slawski} ;  second, we
plan to  derive similar  views on a  wider range  of sparsity-inducing
penalties, such as the fused-Lasso or the OSCAR \citep{Bondell08}.

\bibliographystyle{elsarticle-harv} 
\bibliography{biblio_crafter}

\appendix

\section{Proof of Proposition~\ref{prop:monitoring}}~\label{sec:proof:prop:monitoring}

We detail here a proof yielding a slightly tighter bound. 
Proposition~\ref{prop:monitoring} is simply a corollary of
Proposition~\ref{prop:monitoring:appendix:v2} stated and proved below.

The following Lemma relates the penalty
associated with a infeasible $\bfgamma$-value 
($\norm[*]{\bfgamma} > \eta$) to the one obtained by shrinking this
$\bfgamma$-value to reach the boundary of $\uball[*]^\eta$.

\begin{lemma}\label{prop:lemma1:v2}
  Let $\clS \subseteq \{1,...,p\}$, $\alpha\in(0,1)$, $\norm[*]{\cdot}$ be a vectorial norm and $\varphi^{*}(\cdot,\clS,\alpha) :
  \mathbb{R}^{p} \rightarrow \mathbb{R}^{p}$ be defined as follows:
  \begin{equation*} 
    \left\{
    \begin{array}{l}
      \varphi_{\clS}^{*}(\bfgamma,\clS,\alpha) = \bfgamma_{\clS} \\
      \varphi_{\clS^c}^{*}(\bfgamma,\clS,\alpha) = \alpha \bfgamma_{\clS^c} 
    \end{array}\right.\enspace.
   \end{equation*} 
   Then,
   \begin{equation}\label{eq:lemma1:v2}
     \norm{\bfbeta-\varphi^{*}(\bfgamma,\clS,\alpha)}^2 \geq 
      \alpha \norm{\bfbeta-\bfgamma}^2  -
      \alpha (1-\alpha) \norm{\bfgamma_{\clS^c}}^{2} 
     \enspace. 
  \end{equation}
  \begin{proof}
    \begin{align*}
      \norm{\bfbeta-\varphi^{*}(\bfgamma,\clS,\alpha)}^2 & =
      \norm{\bfbeta_{\clS}-\bfgamma_{\clS}}^2 +
      \alpha \norm{\bfbeta_{\clS^c}-\bfgamma_{\clS^c}}^2 +
      (1-\alpha) \norm{\bfbeta_{\clS^c}}^{2} -
      \alpha (1-\alpha) \norm{\bfgamma_{\clS^c}}^{2} \\
      & \geq
      \norm{\bfbeta_{\clS}-\bfgamma_{\clS}}^2 +
      \alpha \norm{\bfbeta_{\clS^c}-\bfgamma_{\clS^c}}^2  -
      \alpha (1-\alpha) \norm{\bfgamma_{\clS^c}}^{2}  \\
      & \geq
      \alpha \norm{\bfbeta-\bfgamma}^2  -
      \alpha (1-\alpha) \norm{\bfgamma_{\clS^c}}^{2} 
      \enspace.
    \end{align*}
  \end{proof}
\end{lemma}
\begin{proposition}\label{prop:monitoring:appendix:v2}
  For any vectorial norm $\norm[*]{\cdot}$, 
  $\forall \bfgamma \in \mathbb{R}^{p}:\norm[*]{\bfgamma} \geq \eta$, and 
  $\forall (\clS,\alpha)\in2^{\{1,\ldots,p\}}\times(0,1)$ such that 
  $\norm[*]{(\bfgamma_{\clS},\alpha\bfgamma_{\clS^c })}\leq \eta$, we   
  have:
  \begin{equation*}
    \min_{\bfbeta\in\mathbb{R}^{p}} \max_{\bfgamma' \in \uball[*]^\eta} 
    J_\lambda(\bfbeta,\bfgamma') 
    \geq
    \alpha J_\lambda\left(\bfbeta^\star\left(\bfgamma\right),\bfgamma\right) -
    \lambda \alpha (1-\alpha) \norm{\bfgamma_{\clS^c}}^{2}
    \enspace,
  \end{equation*}
  where 
  \begin{equation*}
    J_\lambda(\bfbeta,\bfgamma) = \norm{\bfX \bfbeta - \bfy}^2 + 
      \lambda \norm{\bfbeta - \bfgamma}^2
    \enspace \text{and} \enspace
    \bfbeta^\star(\bfgamma) = \argmin_{\bfbeta\in\mathbb{R}^{p}} J_\lambda(\bfbeta,\bfgamma)
    \enspace.
  \end{equation*}
  \begin{proof} 
  For all $\bfbeta \in \mathbb{R}^{p}$ and for any $(\bfgamma,\clS,\alpha) \in
  \mathbb{R}^{p}\times2^{\{1,\ldots,p\}}\times(0,1)$ such that
  $\varphi^{*}(\bfgamma,\clS,\alpha) \in \uball[*]^\eta$, with $\varphi^{*}$ 
  defined as in Lemma \ref{prop:lemma1:v2} we have
  \begin{equation}\label{eq:maxgreater:v2}
    \max_{\bfgamma' \in \uball[*]^\eta} 
    J_\lambda(\bfbeta,\bfgamma') 
    \geq
    J_\lambda\left(\bfbeta,\varphi^{*}(\bfgamma,\clS,\alpha)\right)
    \enspace,
  \end{equation}
  since $\varphi^{*}(\bfgamma,\clS,\alpha)$ belongs to $\uball[*]^\eta$. 
  We now compute a lower bound of the right-hand-side for $\bfgamma$ such that
  $\norm[*]{\bfgamma} \geq \eta$:
    \begin{align}       
      J_\lambda\left(\bfbeta,\varphi^{*}(\bfgamma,\clS,\alpha)\right) 
        & = \alpha \left( 
              \frac{1}{\alpha} \norm{\bfX \bfbeta - \bfy}^2 + 
              \frac{\lambda}{\alpha} 
              \norm{\bfbeta - \varphi^{*}(\bfgamma,\clS,\alpha)}^2
            \right)  
            \nonumber \\
        & \geq \alpha \left( 
              \norm{\bfX \bfbeta - \bfy}^2 + 
              \frac{\lambda}{\alpha} 
              \norm{\bfbeta - \varphi^{*}(\bfgamma,\clS,\alpha)}^2
            \right)  
            \label{eq:crude_inequality} \\
        & \geq \alpha \left( 
              \norm{\bfX \bfbeta - \bfy}^2 + 
              \lambda \norm{\bfbeta - \bfgamma}^2
           \right) -
           \lambda\alpha (1-\alpha) \norm{\bfgamma_{\clS^c}}^{2} 
           \enspace, \nonumber
    \end{align}
    where the last inequality stems from Lemma~\ref{prop:lemma1:v2}.
    This inequality holds for any given $\bfbeta$-value, in particular for
    $\bfbeta^\star(\varphi^{*}(\bfgamma,\clS,\alpha))=\argmin_{\bfbeta \in \mathbb{R}^{p}}
    J_\lambda\left(\bfbeta,\varphi^{*}(\bfgamma,\clS,\alpha)\right)$:
    \begin{align}       
      \min_{\bfbeta\in\mathbb{R}^{p}} 
      J_\lambda\left(\bfbeta,
                     \varphi^{*}(\bfgamma,\clS,\alpha)
               \right) 
      & \geq \alpha
          J_\lambda\left(\bfbeta^\star\left(\varphi^{*}(\bfgamma,\clS,\alpha)\right)\right) -
          \lambda\alpha (1-\alpha) \norm{\bfgamma_{\clS^c}}^{2} 
          \nonumber \\
      & \geq \alpha
          J_\lambda\left(\bfbeta^\star\left(\bfgamma\right),\bfgamma\right) -
          \lambda\alpha (1-\alpha) \norm{\bfgamma_{\clS^c}}^{2} 
          \label{eq:inequality:particular:appendix:v2} 
          \enspace,
    \end{align}
    where the second inequality follows from the definition of
    $\bfbeta^\star(\bfgamma)$.
    Inequality \eqref{eq:inequality:particular:appendix:v2} can be restated as:
     \begin{align*}       
       \min_{\bfbeta\in\mathbb{R}^{p}} 
       J_\lambda\left(\bfbeta,\varphi^{*}(\bfgamma,\clS,\alpha)\right)
       & \geq
         \alpha \min_{\bfbeta\in\mathbb{R}^{p}} 
         J_\lambda(\bfbeta,\bfgamma)  -
         \lambda\alpha (1-\alpha) \norm{\bfgamma_{\clS^c}}^{2} 
       \enspace.
    \end{align*}
    We finally remark that,  
    since $\varphi^{*}(\bfgamma,\clS,\alpha) \in \uball[*]^\eta$,
    we trivially have:
    \begin{align*}
      \min_{\bfbeta\in\mathbb{R}^{p}} \max_{\bfgamma \in \uball[*]^\eta} J_\lambda(\bfbeta,\bfgamma) 
      & \geq 
       \min_{\bfbeta\in\mathbb{R}^{p}} 
        J_\lambda\left(\bfbeta,\varphi^{*}(\bfgamma,\clS,\alpha)\right)
     \enspace,
    \end{align*}
    which concludes the proof.
  \end{proof}
\end{proposition}
Proposition~\ref{prop:monitoring} follows by choosing $\alpha={\eta}/{\norm[*]{\bfgamma}}$.

\end{document}